\documentclass[a4paper,11pt]{article}
\usepackage{fullpage}
\usepackage{epsf}
\usepackage{fancyhdr}
\usepackage{graphics}
\usepackage{graphicx}
\usepackage{psfrag}
\usepackage[T1]{fontenc}
\usepackage{color}
\usepackage{amsthm}
\usepackage{amsfonts}
\usepackage{amsmath}
\usepackage{amssymb}
\usepackage{mathrsfs}
\usepackage{accents}
\usepackage{listings}
\usepackage{caption}
\usepackage{subcaption}
\usepackage[linesnumbered, ruled, vlined]{algorithm2e}
\usepackage[titletoc,toc,title]{appendix}
\newlength{\widebarargwidth}
\newlength{\widebarargheight}
\newlength{\widebarargdepth}

\makeatletter
\long\def\@makecaption#1#2{
        \vskip 0.8ex
        \setbox\@tempboxa\hbox{\small {\bf #1:} #2}
        \parindent 1.5em  
        \dimen0=\hsize
        \advance\dimen0 by -3em
        \ifdim \wd\@tempboxa >\dimen0
                \hbox to \hsize{
                        \parindent 0em
                        \hfil 
                        \parbox{\dimen0}{\def\baselinestretch{0.96}\small
                                {\bf #1.} #2
                                } 
                        \hfil}
        \else \hbox to \hsize{\hfil \box\@tempboxa \hfil}
        \fi
        }
\makeatother

\usepackage{amssymb}
\usepackage{amsmath}
\usepackage{amsthm}
\usepackage{enumerate}
\usepackage{verbatim} 
\usepackage[titletoc,toc,title]{appendix}
\usepackage{csquotes} 
\usepackage{upquote} 
\usepackage[bottom]{footmisc} 
\usepackage{graphicx} 
\usepackage[ruled]{algorithm2e}
\usepackage{thmtools}
\usepackage{thm-restate}
\usepackage[dvipsnames]{xcolor}
\usepackage[colorlinks,linkcolor = RawSienna, urlcolor  = violet, citecolor = RoyalBlue, anchorcolor = violet]{hyperref}
\usepackage{cleveref}

\usepackage[english]{babel} 
\renewcommand{\baselinestretch}{1.04} 
\date{}
\usepackage[style = alphabetic,citestyle=alphabetic,maxbibnames=99,backend=biber,natbib=true,backref=true]{biblatex}
\DefineBibliographyStrings{english}{%
  backrefpage = {Cited on page},
  backrefpages = {Cited on pages},
}

\newcommand*\tr[0]{\text{tr}}

\newcommand\numberthis{\addtocounter{equation}{1}\tag{\theequation}}

\def\lv{\lVert}
\def\rv{\rVert}

\newcommand{\BlackBox}{\rule{1.5ex}{1.5ex}}  
\newenvironment{Proof}{\par\noindent{\bf Proof\ }}{\hfill\BlackBox\\[2mm]}
\newtheorem{theorem}{Theorem}

\newtheorem{lemma}[theorem]{Lemma}

\newcommand{\argmax}{\operatornamewithlimits{argmax}}

\newcommand\psued[1]{R^s_{#1}}

\newcommand\regret[1]{R^c_{#1}}

\newcommand\algname{\mathsf{OSOM}}
\newcommand\UCB{\mathsf{UCB}}
\newcommand\OFUL{\mathsf{OFUL}}
\newcommand\corral{\mathsf{CORRAL}}
\newcommand\moss{\mathsf{MOSS}}
\newcommand\linucb{\mathsf{LinUCB}}
\newcommand\expalg{\mathsf{EXP}4}

\newcommand\sao{\mathsf{SAO}}

\newcommand\tstar{\tau_*}
\usepackage[shortlabels]{enumitem}
\newcommand\unitball{\mathbb{B}_2^d(1)}

\newcommand\mustar[1]{\mu_{#1}}

\newcommand\thetastar{\theta^{*}}

\newcommand\iinrange[2]{i \in \{#1,\ldots, #2\}}

\newcommand\alpt[2]{\alpha_{#1,#2}}

\newcommand\muit[2]{\tilde{\mu}_{#1,#2}}

\newcommand\git{g_{i,t}}

\newcommand\ccomplex{\mathcal{C}_t^c}

\newcommand\Tstar{\theta^*}

\newcommand\That{\hat{\theta}_t}

\newcommand\Aboldt{\boldsymbol{\alpha}_{K+1:t}}
\newcommand\Gboldt{\mathbf{G}_{K+1:t}}

\newcommand\Aboldtn{\boldsymbol{\alpha}}
\newcommand\Gboldtn{\mathbf{G}}
\newcommand\etabold{\boldsymbol{\eta}}
\newcommand\mubold{\boldsymbol{\mu}}
\newcommand\mutbold{\boldsymbol{\tilde{\mu}}}

\newcommand\averagerew{\bar{g}_{i,t}}

\newcommand\tmin{\tau_{\min}(\delta',n)}
\newcommand\updela{\Upsilon_{\delta'}(t,n)}
\newcommand\mone{\mathcal{M}_{\delta'}(t)}

\newcommand\kenv{\mathcal{K}_{\delta'}(t,n)}
\newcommand\kenvs{\mathcal{K}_{\delta'}(s-1,n)}
\newcommand\kenvt{\mathcal{K}_{\delta'}(t,n)}
\newcommand\wenv{\mathcal{W}_{\delta'}(t-1,n)}
\newcommand\wenvone{\mathcal{W}_{\delta'}(t,n)}
\newcommand\wenvtau{\mathcal{W}_{\delta'}(\tstar-1,n)}
\newcommand\wenvn{\mathcal{W}_{\delta'}(n,n)}

\newcommand\loweig{\rho_{\min}}
\newcommand\higheig{\rho_{\max}}

\usepackage{mathtools}

\DeclareMathSizes{9}{8}{7}{5}
\title{\textbf{\textsc{Osom}: A simultaneously optimal algorithm for multi-armed and linear contextual bandits}}
\author{
Niladri S. Chatterji \quad
 Vidya Muthukumar \quad
Peter L. Bartlett
\\
\textsf{\{chatterji,vidya.muthukumar,peter\}@berkeley.edu} 
\\ \\
University of California, Berkeley
}
\date{\today}
\begin{document}

\maketitle
\begin{abstract}
    We consider the stochastic linear (multi-armed) contextual bandit problem with the possibility of hidden \textit{simple multi-armed bandit} structure in which the rewards are independent of the contextual information.
    Algorithms that are designed solely for one of the regimes are known to be sub-optimal for the alternate regime.
    We design a single computationally efficient algorithm that simultaneously obtains problem-dependent optimal regret rates in the simple multi-armed bandit regime and minimax optimal regret rates in the linear contextual bandit regime, without knowing a priori which of the two models generates the rewards.
    These results are proved under the condition of stochasticity of contextual information over multiple rounds.
    Our results should be viewed as a step towards principled data-dependent policy class selection for contextual bandits.
\end{abstract}
\section{Introduction}

The \textit{contextual bandit} paradigm involves sequential decision-making settings in which we repeatedly pick one out of $K$ actions (or ``arms'') in the presence of contextual side information.
Algorithms for this problem usually involve policies that map the contextual information to a chosen action, and the reward feedback is typically \textit{limited} in the sense that it is only obtained for the action that was chosen.
The goal is to maximize the total reward over several ($n$) rounds of decision-making, and the performance of an online algorithm is typically measured in terms of \textit{regret} with respect to the best policy within some policy class $\Pi$ that is fixed a priori.
Applications of this paradigm include advertisement placement/web article recommendation~\cite{li2010contextual, agarwal2016making}, clinical trials and mobile health-care~\cite{woodroofe1979one,tewari2017ads}.

The contextual bandit problem can be thought of as an online supervised learning problem (over policies mapping contexts to actions) with limited information feedback, and so the optimal regret bounds scale like $\mathcal{O}(\sqrt{K n \log |\Pi|})$, a natural measure of the sample complexity of the policy class~\cite{auer2002nonstochastic,mcmahan2009tighter,beygelzimer2011contextual}.
These are typically achieved by algorithms that are inefficient (linear in the size of the policy class).
Much of the research in contextual bandits has tackled computational efficiency~\cite{langford2008epoch,agarwal2014taming,rakhlin2016bistro,syrgkanis2016efficient,syrgkanis2016improved, foster2018contextual}: do there exist computationally efficient algorithms that achieve the optimal regret guarantee?
A question that has received relatively less attention involves the choice of policy class itself.
Even for a fixed regret-minimizing algorithm, the choice of policy class is critical to maximize the overall \textit{reward} of the algorithm.
As can be seen in applications of contextual bandits models for article recommendation~\cite{li2010contextual}, the choice is often made in hindsight, and more complex policy classes are used if the algorithm is run for more rounds.
A quantitative understanding of how to do this is still lacking, and intuitively, we should expect the optimal choice of policy class to not be static.
Ideally, we could design adaptive contextual bandit algorithms that would initially use simple policies, and switch over to more complex ones as more data is obtained.

Theoretically, what this means is that the regret bounds derived for a contextual bandit algorithm are only meaningful for rewards that are generated by a policy within the policy class to which the algorithm is tailored.
If the rewards are derived from a ``more complex" policy outside the policy class, even the optimal policy may neglect obvious patterns and obtain a very low reward.
If the rewards are derived from a policy that is expressible by a much smaller class, the regret that is accumulated is unnecessary.
Let us view this through the lens of the simplest possible example: the standard linear contextual bandits~\cite{chu2011contextual} paradigm, where we can choose one out of $K$ arms and rewards are generated according to the process
\begin{align*}
    g_{i,t} = \mu_i + \langle \thetastar, \alpt{i}{t}\rangle + \eta_{i,t}, \text{ for all } i \in [K],
\end{align*}
where $\mustar{i}$ represents a ``bias'' of arm $i$, $\thetastar \in \mathbb{R}^d$ represents the linear parameter of the model (which is shared across all arms\footnote{This is the model that was described in~\cite{chu2011contextual}. It is worth noting that more complex variants of this model with a separate $\theta^*_i$ for every $i \in [K]$ have also been empirically evaluated~\cite{li2010contextual}.
}), $\alpt{i}{t} \in \mathbb{R}^d$ represents the contextual information and $\{\eta_{i,t}\}_{t=1}^n$ represents noise in the reward observations.
It is well-known that variants of linear upper confidence bound algorithms like $\linucb$ \citep{chu2011contextual} and $\OFUL$ \cite{abbasi2011improved}\footnote{Guarantees for $\OFUL$ were established under slightly different constraints on $\thetastar$ and the context vectors which led to a regret bound of~$\widetilde{\mathcal{O}}((d+K)\sqrt{n})$. We show in Lemma \ref{lem:regretboundafterfailure} that a slight variant of $\OFUL$ has its regret bounded by~$\widetilde{\mathcal{O}}((\sqrt{d}+\sqrt{K})\sqrt{n})$ in our setting.} suffer at most $\widetilde{\mathcal{O}}((\sqrt{d} + \sqrt{K})\sqrt{n})$ regret with respect to the optimal linear policy (where $\widetilde{\mathcal{O}}$ is up-to logarithmic factors in any of $(d,K,n)$).
However, setting $\thetastar = 0$ yields the important case of the reward distribution being independent from the contextual information.
Here, a simple upper confidence bound algorithm like $\UCB$ \citep{auer2002finite} would yield the optimal $\mathcal{O}(\log n)$ regret bound, \textit{which does not depend on the dimension of the contexts $d$}.
Thus, we pay substantial extra regret by using the algorithm meant for linear contextual bandits on such instances with much simpler structure.
On the other hand, upper confidence bounds that ignore the contextual information will not guarantee any control on the policy regret: it can even be linear.
It is natural to desire a single approach that adapts to the inherent complexity of the reward-generating model and obtains the optimal regret bound as if this complexity was known in hindsight.
Specifically, this paper seeks an answer to the following question:

\emph{Does there exist a single algorithm that simultaneously achieves the $\mathcal{O}(\log n)$ regret rate on simple multi-armed bandit instances and the $\widetilde{\mathcal{O}}((\sqrt{d}+\sqrt{K})\sqrt{n})$ regret rate on linear contextual bandit instances?}

\subsection{Our contributions}

    We answer the question of simultaneously optimal regret rates in the multi-armed (``simple'') bandit regime and the linear contextual (``complex'') bandit regime affirmatively under the condition that the contexts are generated from a stochastic process that yields covariates that are not ill-conditioned. 
    Our algorithm, $\algname$ (for Optimistic Selection of Models), essentially exploits the best policy (simply the best arm) that is learned under the assumption of the simple reward model - while conducting a sequential statistical test for the presence of additional complexity in the model, and particularly \textit{whether ignoring this additional complexity would lead to substantial regret}.
    This is a simple statistical principle that could conceivably be generalized to arbitrary policy classes \textit{that are nested}: we will see that the $\algname$ algorithm critically exploits the nested structure of the simple bandit model within the linear contextual model.

\subsection{Related work}

The contextual bandit paradigm was first considered by \citeauthor{woodroofe1979one} (\citeyear{woodroofe1979one}) to model clinical trials. Since then it has been studied intensely both theoretically and empirically in many different application areas under many different pseudonyms. We point the reader to \citep{tewari2017ads} for an extensive survey of the contextual bandits history and literature. 

Treating policies as experts ($\expalg$ \citep{auer2002nonstochastic}) with careful control on the exploration distribution led to the optimal regret bounds of $\mathcal{O}(\sqrt{K n \log |\Pi|})$ in a number of settings.
From an \textit{efficiency} point of view (where efficiency is defined with respect to an \textit{arg-max-oracle} that is able to compute the best greedy policy in hindsight), the first approach conceived was the epoch-greedy approach~\cite{langford2008epoch}, that suffers a sub-optimal dependence of $n^{2/3}$ in the regret. More recently, ``randomized-UCB" style approaches~\cite{agarwal2014taming} have been conceived that retain the optimal regret guarantee with $\widetilde{\mathcal{O}}(\sqrt{n})$ calls to the arg-max-oracle. This question of computational efficiency has generated a lot of research interest~\cite{rakhlin2016bistro,syrgkanis2016efficient,syrgkanis2016improved, foster2018contextual}.
The problem of policy class selection itself has received less attention in the research community, and how this is done in practice in a statistically sound manner remains unclear. An application of \textit{linear} contextual bandits was to personalized article recommendation using hand-crafted features of users~\cite{li2010contextual}: two classes of linear contextual bandit models with varying levels of complexity were compared to simple (multi-armed) bandit algorithms in terms of \textit{overall reward} (which in this application represented the click-through rate of ads).
A striking observation was that the more complex models won out when the algorithm was run for a longer period of time (eg: 1 day as opposed to half a day).
Surveys on contextual bandits as applied to mobile health-care~\cite{tewari2017ads} have expressed a desire for algorithms that adapt their choice of policy class according to the amount of information they have received (e.g. the number of rounds).
At a high level, we seek a theoretically principled way of doing this.

Perhaps the most relevant work to online policy class selection involves significant attempts to \textit{corral} a band of $M$ base bandit algorithms into a meta-bandit framework~\cite{agarwal2017corralling}.
The idea is to bound the regret of the meta-algorithm in terms of the regret of the best base algorithm in hindsight. (This is clearly useful for policy class selection that we study here -- by corralling together an algorithm designed for the linear model and one for the simple multi-armed bandits model.)
The Corral framework is very general and can be applied to any set of base algorithms, whether efficient or not.
This generality is attractive, but it is not the optimal choice of \textit{computationally efficient} algorithm for the multi-armed-vs-linear-contextual bandit problem for a couple of reasons. \begin{enumerate}
    \item It is not clear what (if any) choice of base algorithms would lead to a computationally efficient algorithm that is also statistically optimal in a minimax sense simultaneously for both problems.
    \item The meta-algorithm framework uses an experts algorithm (in particular, mirror descent with log-barrier regularizer and importance weighting on the base algorithms) to choose which base algorithm to play in each round.
    Thus, it is impossible to expect the instance-optimal regret rate of $\mathcal{O}(\log n)$ on the simple bandit instance.
    More generally, the Corral framework will not yield instance-optimal rates on any policy class\footnote{On our much simpler instance of bandit-vs-linear-bandit, we do obtain instance-optimal rates for at least the simple bandit model.}.
\end{enumerate} 

The Corral framework highlights the principal difficulty in contextual bandit model selection that can be thought of as an even finer exploration-exploitation tradeoff: algorithms (designed for particular model classes) that fall out of favor in initial rounds could be picked very rarely and the information required to truly perform model selection may be absent even after many rounds of play.
$\corral$ tackles this difficulty using the log-barrier regularizer for the meta-algorithm as a natural form of heightened exploration~\cite{foster2016learning}, together with clever learning rate schedules. 

Closely related is the concurrent work of~\citep{foster2019model} which tackles the problem of selecting among a hierarchy of linear classes with growing dimension. They work with stochasticity assumptions on the contexts that are \emph{weaker} than the assumptions that we make in our paper. However, they are only able to establish a sub-optimal bound on the regret of~$\widetilde{\mathcal{O}}(d_*^{1/3}n^{2/3})$ (where $d_*$ is dimension of the optimal linear policy) as opposed to the minimax optimal regret rates (that scale with~$n^{1/2}$) which we establish in our paper.

Our stylistic approach to the model selection problem is a little different, as we focus on the much more specific case of $2$ models: the simple multi-armed bandit model and the linear contextual bandit model.
We encounter a similar difficulty and obtain striking clarity on the extent of this difficulty owing to the simplicity of the models.
On the other hand, we observe that commonly encountered sequences of contexts can help us carefully navigate the finer exploration-exploitation tradeoff when the model classes are nested.

Our algorithm ($\algname$) utilizes a simple ``best-of-both-worlds'' principle: exploit the possible simple reward structure in the model until (unless) there is significant statistical evidence for the presence of complex reward structure \textit{that would incur substantial complex policy regret if not exploited}.
This algorithmic framework is inspired by the initial ``best-of-both-worlds'' results for stochastic and adversarial multi-armed bandits; in particular, the ``Stochastic and Adversarial Optimal'' ($\sao$) algorithm~\cite{bubeck2012best} (although the details of the phases of the algorithm and the statistical test are very different).
In that framework, instances that are not stochastic (and could be thought of as ``adversarial'') are not always detected as such by the test.
The test is designed in an elegant manner such that the regret is optimally bounded on instances that are not detected as adversarial, \textit{even if an algorithm meant for stochastic rewards is used.}
Our test to distinguish between simple and complex instances shares this flavor -- in fact, not all theoretically complex instances ($\theta^* \neq 0$) are detected as such. 

Also related are results on contextual bandits with similarity information \textit{on the contexts}, which automatically encodes a potentially easier learning problem~\cite{slivkins2014contextual}.
The main novelty in these results involves adapting to such similarity online.

Technically, our proofs leverage 
recent 
theoretical results on regret bounds for linear bandits~\cite{abbasi2011improved}, which can easily be applied to the linear contextual bandit model, and sophisticated \textit{self-normalized} concentration bounds for our estimates of both the bias terms $\mu_i$ and the parameter vector $\theta^*$.
For the latter, we find that the Matrix Freedman inequality~\cite{oliveira2009concentration,tropp2011freedman} is particularly useful.

\section{Problem statement}

\paragraph{Notation and definitions.} Given a vector $v$, let $v_i$ denote its $i^{th}$ component. For a vector we let $\lv v \rv_{p}$ for $p\in [1,\infty]$ denote the $\ell_p$-norm. Given a matrix $M$ we denote it's operator norm by $\lv M \rv_{op}$, and use $\lv M \rv_{F}$ to denote its Frobenius norm. Given a symmetric matrix $S$ let $\gamma_{\max}(S)$ and $\gamma_{\min}(S)$ denote its largest and smallest eigenvalues. Given a positive definite matrix $V$ we define the norm of a vector $w$ with respect to matrix $V$ as $\lv w \rv_V^2 = w^{\top}V w$. 

A stochastic process $\{\xi_t\}_{t=1}^{\infty}$, defined with respect to a filtration $\{\mathcal{H}_t\}_{t=1}^{\infty}$, is said to be conditionally $\sigma$-sub-Gaussian for some $\sigma >0$ if, for all $\lambda \in \mathbb{R}$, we have 
\begin{align*}
\mathbb{E}\left[e^{\lambda \xi_t} \big \lvert \mathcal{H}_{t-1}\right] \le \exp(\lambda^2 \sigma^2/2).
\end{align*}

\subsection{Setup} 
At the beginning of each round $t \in [n]$, the learner is required to choose one of $K$ arms and gets a \emph{reward} associated with that arm. To help make this choice the learner is handed a context vector at every round $\alpha_t = [\alpha_{1,t},\ldots,\alpha_{K,t}] \in \mathbb{R}^{d\times K}$ (this is essentially a concatenation of $K$ vectors, each of dimension equal to $d$). Let $\git$ denote the reward of arm $i$ and let $A_t \in [K]$ denote the choice of the learner in round $t$. The rewards could be arriving from one of two models that is described below:

\textbf{Simple Model}: Under the simple \textit{multi-armed} bandit model, the mean rewards of $K$ arms are fixed and are \emph{not} a function of the contexts. That is, at each round
\begin{align*}
    g_{i,t} = \mustar{i} + \eta_{i,t},\qquad  \text{ for all } i\in [K]
    \end{align*}
where $\mu_i \in [-1,1]$, $\{\eta_{i,t}\}_{i=1}^K$ are identical, independent, zero mean, $\sigma$-sub-Gaussian noise (defined below). Let the arm with the highest reward have mean $\mu^*$ and be indexed by $i^*$. The benchmark that the algorithm hopes to compete against is the \emph{pseudo-regret} (henceforth \emph{regret} for brevity),
\begin{align*}
    \psued{n} := n \mu^* -\sum_{s = 1}^n \mustar{A_s} .
\end{align*} 
Define the gap as the difference in the mean reward of the best arm compared to the mean reward of the $i^{th}$ arm, that is, $\Delta_i := \mu^* - \mu_i $. Previous literature on multi-armed bandits \citep{Lai:1985:AEA:2609660.2609757} tells us that the best one can hope to do in this setting in the worst case is $\mathbb{E}\left[\psued{n}\right] = \Omega(\sum_{i}\log(n)/\Delta_i)$. Several algorithms like $\UCB$ \citep{auer2002finite} and  $\moss$ \citep{audibert2010regret,degenne2016anytime} achieve this lower bound up to logarithmic 
factors. 

\textbf{Complex Model}: In this model the mean reward of each arm is a linear  function of the contexts (linear contextual bandits). We work with the following stochastic assumptions on the context vectors. 

We define the filtration $\mathcal{F}_{t-1}$ to be the $\sigma$-algebra generated by all previous contexts, rewards seen, and choices of the algorithm up until the end of round $t-1$; that is,  $\mathcal{F}_{t-1} := \{\alpha_{s},\{\eta_{i,s}\}_{i\in [K]},A_s,g_{A_s},i_s,j_s\}_{s\in [t-1]}$ (where $i_s$ and $j_s$ are defined in equations \eqref{eq:simpleestimate} and \eqref{eq:complexestimate} respectively), and throughout this paper, we denote as shorthand $\mathbb{E}_{t-1}\left[\cdot\right]$ as the conditional expectation with respect to this filtration.
Then, we assume that each of these context vectors $\alpt{i}{t} \in \unitball$ are drawn independent of the past from a distribution such that $\alpha_{i,t}$ is independent of $\{\alpha_{j,t}\}_{j\neq i}$ and, $\forall \; i \in [K]$ and $\forall \; t\in [n]$,
\begin{align}
  \nonumber &\mathbb{E}_{t-1}\left[\alpha_{i,t}\right] = 0, \\ &\mathbb{E}_{t-1}\left[\alpha_{i,t}\alpha_{i,t}^{\top}\right] = \Sigma_c \succeq \loweig \cdot I \succ 0 , \label{def:assumptioncontext}
\end{align}
That is, the conditional mean of the context vectors are $0$ and the covariance matrix has its minimum eigenvalue \textit{bounded below} by $\loweig$. 
Note that because we have assumed bounded contexts, we have $\loweig \leq 1/d$ and so tracking the dependence of our regret bounds on $\loweig$ will be important.
{Also note that the above assumption includes, as a special case, contexts that evolve according to a stochastic process exogeneous to the algorithmic actions and previously observed rewards.}

Further, the constraint $\alpha_{i,t} \in \unitball$ automatically implies that each of the context vectors $\{\alpha_{i,t}\}_{i \in [K], t \in [n]}$ are conditionally sub-Gaussian with parameter $\higheig \leq 1$.
We will see that our regret bounds will depend on the ratio $\higheig/\loweig$, which we will call a ``sub-Gaussian condition number".
In general, for sufficiently \textit{diverse} contexts this condition number will be bounded by a dimension-free constant; assumptions similar to this have been made in past work analyzing the greedy algorithm for linear contextual bandits~\citep{bastani2017mostly,kannan2018smoothed,raghavan2018externalities}.

We note that the commonly used algorithms for the linear contextual bandit problem like $\OFUL$ work even with \emph{adversarially} distributed contexts and stochastic conditional rewards.
This is significantly more general than the set of stochastic assumptions that we make here. 
However, our goal is to select the right model \emph{optimally} under these stochastic assumptions on the contexts between the simple model and the complex model, which previous algorithms like $\OFUL$ would fail to do.
Understanding the possibilities and limits of model selection in the case of adversarial contexts is an interesting direction for future work.

In this complex model, we assume there exists an underlying linear predictor $\thetastar \in \mathbb{R}^d$ and \textit{biases} $[  \mustar{1},\ldots,\mustar{K} ] \in \mathbb{R}^{K}$ of the $K$ arms, such that the mean rewards of the arms are affine functions of the contexts, that is,
\begin{align*}
    g_{i,t} =  \mustar{i} + \langle \thetastar, \alpt{i}{t}\rangle + \eta_{i,t} .
\end{align*}
We impose boundedness constraints on the parameters: in particular, we have $\mustar{i} \in [-1,1]$, $\theta^* \in \unitball$.
Further, the noise $\{\eta_{i,t}\}_{t=1}^n$ are identical, independent, zero mean, and $\sigma$-sub-Gaussian. 
Clearly, simple model instances (which are parameterized only by the biases $[  \mustar{1},\ldots,\mustar{K} ] \in \mathbb{R}^{K}$) can be expressed as complex model instances by setting $\thetastar = 0$.

At each round define $\kappa_t=\argmax_{\kappa\in \{1,\ldots,K\}_{i=1}^K}\left\{ \mu_{\kappa}+\langle \theta^*,\alpha_{\kappa,t}\rangle \right\}$ to be the best arm at round $t$. Here, we define pseudo-regret with respect to the optimal policy under the generative linear model:
\begin{align*}
   \regret{n} :=\sum_{s=1}^n \left[\mu_{\kappa_s}+\langle \theta^*,\alpha_{\kappa_s,s}\rangle - \mu_{A_s}-\langle \theta^*,\alpha_{A_s,s}\rangle \right].
\end{align*}
As noted above, past literature on this problem yielded algorithms like $\linucb$~\citep{chu2011contextual} and $\OFUL$~\citep{abbasi2011improved} that only suffer from the minimax regret of $\widetilde{\mathcal{O}}((\sqrt{d}+\sqrt{K})\sqrt{n})$. As we will see in the simulations, these algorithms actually incur the dependence on the dimension in the regret, even for simple instances.
\section{Construction of Confidence Sets}
In our algorithm, which is presented in Section~\ref{section:algorithm}, at the \emph{end of round} $t$, we build an upper confidence estimate for each arm. Let $T_i(t): = \sum_{s=1}^t \mathbb{I}\left[A_s = i\right]$ be the number of times arm $i$ was pulled and $\averagerew := \sum_{s=1}^t g_{i,s}\mathbb{I}\left[A_s = i\right]/T_i(t)$ be the average reward of that arm at the end of round $t$. For each arm we define the upper confidence estimate for any $\delta' > 0$:
\begin{align}
    &\label{def:ucbarms}
    \muit{i}{t} : = \averagerew + \sigma\left[\frac{1+T_{i}(t)}{T_{i}^2(t)}\left(1+ 2\log\left(\frac{K(1+T_i(t))^{\frac{1}{2}}}{\delta'}\right)\right)\right]^{\frac{1}{2}}.
\end{align}
  Lemma~6 in \citep{abbasi2011improved} (restated below as Lemma~\ref{lem:csimplebound} here) uses a refined self-normalized martingale concentration inequality to bound $\lvert \mu_i - \averagerew \rvert$ across all arms and all rounds. 

\begin{lemma}\label{lem:csimplebound} Under the simple model, with probability at least $1-\delta'$ we have, for all $i \in \{1,\ldots,K\}$ and for all $t\ge 0$,
\begin{align*}
     \lvert \mu_i - \averagerew \rvert\le \sigma\left[\frac{1+T_{i}(t)}{T_{i}^2(t)}\left(1+ 2\log\left(\frac{K(1+T_i(t))^{\frac{1}{2}}}{\delta'}\right)\right)\right]^{\frac{1}{2}}.
\end{align*}
\end{lemma}
For any round $t>K$, let $\That$ be the $\ell^2$-regularized least-squares estimate of $\Tstar$ defined below.
\begin{align}
\label{def:ohat}
\That = \left(\Aboldt^{\top}\Aboldt +  I\right)^{-1} \Aboldt^{\top} \Gboldt,
\end{align}
where $\Aboldt$ is the matrix whose rows are the context vectors selected from round $K+1$ up until round $t$: $\alpha_{A_{K+1},K+1}^{\top},\ldots,\alpha_{A_t,t}^{\top}$ and $\Gboldt = [g_{A_{K+1},K+1}- \tilde{\mu}_{A_{K+1},K},\ldots,g_{A_t,t}- \muit{A_t}{t-1}]^{\top}$. Here we are regressing on the rewards seen to estimate $\theta^*$, while using the bias estimates $\muit{i}{t-1}$ obtained by our upper confidence estimates defined in Eq.~\eqref{def:ucbarms}.
\begin{restatable}{lemma}{thetaset}\label{prop:thetaconfidence} Let $\That$ be defined as in Eq.~\eqref{def:ohat}. Then, with probability at least $1-3\delta'$ we have that for all $t>K$, $\theta^*$ lies in the set
\begin{align}
\mathcal{C}_{t}^c &: = \left\{\theta\in \mathbb{R}^d: \lv \theta - \That\rv_{2} \le \kenvt \right\},
 \label{def:ccomplex}
\end{align}
where $\kenvt = \widetilde{\mathcal{O}}\left(\sigma \sqrt{\frac{d}{t \loweig}}\right)$ is defined in Eq.~\eqref{def:kenv}.
\end{restatable}
We prove this lemma in Appendix~\ref{app:omittedproofs}.
\section{Algorithm and main result}\label{section:algorithm} 
\begin{algorithm}[t]
\caption{$\algname$ - Optimistic Selection Of Models \label{algorithm:OFULER}}

\For{$t=1,\ldots,K$}{
Play arm $t$ and receive reward $g_{t,t}$, \qquad \emph{\text{(Play each arm at least once.)}}
}
\For{$t=K+1,\ldots,n$}{
    ${\texttt{Current Model}} \leftarrow \text{`Simple'}$
     
    \textit{Simple Model Estimate: } \begin{align}\label{eq:simpleestimate}
        i_t \in \argmax_{\iinrange{1}{K}} \left\{\tilde{\mu}_{i,t-1}\right\}
    \end{align} \\
    
    \textit{Complex Model Estimate:
    }
    \begin{align}\label{eq:complexestimate}
        j_t,\tilde{\theta}_t \in \argmax_{\iinrange{1}{K}, \theta \in \mathcal{C}_{t-1}^c \cap \mathbb{B}^d_2(1)} \left\{\tilde{\mu}_{i,t-1}+\langle \alpha_{i,t},\theta \rangle\right\}, 
    \end{align}\\
    where $\mathcal{C}_{t-1}^c$ defined in Eq.~\eqref{def:ccomplex}.

    \If{${\emph{\texttt{Current Model}}} = \text{\rm{`Simple'}}$ and $t> K+1$} {Check the condition: \begin{align} \label{test:complextest} \sum_{s=K+1}^{t-1} \left\{\muit{j_{s}}{s-1} + \langle \alpha_{j_s,s},\tilde{\theta}_s\rangle -  g_{i_{s},s}\right\} \le \wenv, \end{align}
    where $\wenvone$ defined in Eq.~\eqref{def:wenv}.
    
    If violated then set ${\texttt{Current Model}} \leftarrow \text{`Complex'}$.
    }
    
    If ${\texttt{Current Model}}=\text{`Simple'}$: Play arm $i_t$ and receive reward $g_{i_t,t}$. \\
    
    Else if ${\texttt{Current Model}}=\text{`Complex'}$: Play arm $j_t$ and receive $g_{j_t,t}.$
    
    Update $\left\{\tilde{\mu}_{i,t}\right\}_{i=1}^K$ and $\ccomplex$.
        
    }    
\end{algorithm}
The intuition behind Algorithm \ref{algorithm:OFULER} is straightforward. The algorithm starts off by using the simple model estimate of the recommended action, that is, $i_t$; until it has reason to believe that there is a benefit from switching to the complex model estimates. If the rewards are truly coming from the simple model, \textit{or from a complex model that is well approximated by a simple multi-armed bandit model}, then Condition \ref{test:complextest} \emph{will not be violated} and the regret shall continue to be bounded under either model. However, if Condition~\ref{test:complextest} \emph{is violated} then algorithm switches to the complex estimates $j_t$ for the remaining rounds. The condition is designed using the function $\wenvone$ which is of the order $\widetilde{\mathcal{O}}(\sigma(\sqrt{d}+\sqrt{K})\sqrt{t})$. This corresponds to the additional regret incurred when we attempt to estimate the extra parameter $\tilde{\theta}_t \in \mathbb{R}^d$. 

At each round Condition~\ref{test:complextest} compares the algorithm's \emph{estimate} for the cumulative reward that could be obtained by playing according to the complex estimates: $\sum_{s=K+1}^{t-1} \muit{j_{s}}{s-1} + \langle \alpha_{j_s,s},\tilde{\theta}_s\rangle$ with the actual cumulative rewards seen so far $\sum_{s=K+1}^{t-1}  g_{i_{s},s}$ by sticking to the simple estimates. 

Under the simple model, given our construction of the confidence sets, the term \newline $\sum_{s=K+1}^{t-1} \langle \alpha_{j_s,s},\tilde{\theta}_s\rangle$ will be bounded by $\widetilde{\mathcal{O}}((\sqrt{d}+\sqrt{K})\sqrt{t})$ as the true underlying vector $\theta^* = 0$, while the remaining terms $\sum_{s=K+1}^{t-1} \muit{j_{s}}{s-1} - g_{i_{s},s}$ shall be at most $\widetilde{\mathcal{O}}(\sqrt{Kt})$, as the simple estimates ($i_s$) shall be picking out the best arm quite often under the simple model. In fact under this model we show in Lemma~\ref{lem:conditionnotviolatedundersimple} that Condition~\ref{test:complextest} is \emph{not violated} with high probability and the algorithm shall continue using simple estimates throughout its entire run.

On the other hand, under the complex model, we switch to the complex estimates only if the difference between the algorithm's estimate for the cumulative reward that could be obtained by playing according to the complex estimates, $\sum_{s=K+1}^{t-1} \muit{j_{s}}{s-1} + \langle \alpha_{j_s,s},\tilde{\theta}_s\rangle$ exceeds the rewards seen so far $\sum_{s=K+1}^{t-1}  g_{i_{s},s}$ by  $\widetilde{\mathcal{O}}((\sqrt{d}+\sqrt{K})\sqrt{t})$. That is, only when the algorithm starts to suffer a regret that is equal to the minimax rate of regret under the complex model. 

While instead if this condition is not violated under the complex model, that is, our estimated cumulative reward for switching to the complex model is close to the rewards seen far. Then we show that the regret under the complex model is small even by using simple estimates. We do this in Lemma~\ref{lem:complexmodeltestnotviolated}.

By combining the arguments outlined above our main theorem optimally bounds the regret of $\algname$ under either of the two reward-generating models.
\begin{theorem}\label{thm:mainregretbound}
With probability at least $1 - 33\delta$, we obtain the following upper bounds on regret for the algorithm $\algname$ (Algorithm~\ref{algorithm:OFULER}):
\begin{enumerate}[label=(\alph*)]
    \item Under the \emph{Simple Model} 
    \begin{align}\label{eq:thmsm}
     \psued{n} \le \sigma\cdot \sum_{i:\Delta_i >0} \left[3\Delta_i + \frac{16}{\Delta_i}\log\left(\frac{2Kn}{\Delta_i \delta}\right)\right].
    \end{align}
    \item Under the \emph{Complex Model} 
    \begin{align}\label{eq:thmcm}
        \regret{n} \le 4(K+1) + 4 \mathcal{W}_{\delta/n}(n,n) = \widetilde{\mathcal{O}}\left\{\sigma\left(\sqrt{\frac{ \higheig}{\loweig}} \cdot \sqrt{d}+\sqrt{K}\right)\sqrt{n}\right\}.
    \end{align}
\end{enumerate}
where $\mathcal{W}_{\delta/n}(n,n)$ is defined in Eq. \eqref{def:wenv}.
\end{theorem}
Notice that Theorem \ref{thm:mainregretbound} establishes regret bounds on the algorithm $\algname$ which are minimax optimal under both \emph{simple model} and the \emph{complex model} up to logarithmic factors. In fact, under the simple model we obtain the optimal \emph{problem-dependent} regret rate.

Under the complex model, Equation~\eqref{eq:thmcm} matches the \textit{minimax-optimal} rate of $\OFUL$ when the ``sub-Gaussian condition number'' of the contexts, $\higheig/\loweig$, is bounded above by a dimension-free constant.
These constitute contexts that are sufficiently \textit{diverse} in their distribution.
An intuitive example of diverse context distributions is the case of the context vectors being uniformly distributed on the discrete hyper-cube\footnote{In a generalization of this example, it would suffice for the entries to be independently sub-Gaussian with parameter $\overline{c}/d$ and each variance lower bounded by $\underline{c}/d$, where $(\overline{c},\underline{c})$ are dimension-free constants.
In this general example, the sub-Gaussian condition number can be verified to be upper bounded by $\overline{c}/\underline{c}$.
} $\{-1/\sqrt{d},1/\sqrt{d}\}^d$, i.e. for every $i \in [K]$ and $t \in [n]$, the $(\alpha_{i,t})_j$ are \text{ i.i.d. } uniform on $\{-1/\sqrt{d},1/\sqrt{d}\}$.
Here, we will have $\higheig = \loweig = 1/d$ and the sub-Gaussian condition number is exactly equal to $1$.
It is important to note that the sub-Gaussian condition number can in general be greater than the actual condition number of the covariance matrix $\Sigma_c$; however, these will be within constant factors of one another for well-behaved distributions on contexts.

Equation~\eqref{eq:thmcm} also provides non-trivial regret rates for the complex model under less friendly context distributions for which the sub-Gaussian condition number could scale with the dimension $d$, but these rates are no longer optimal in the dimension dependence.
Another natural question for future work is whether it is also possible to obtain problem dependent rates in the complex model simultaneously. For example under the complex model by using $\OFUL$ it is possible to show that regret grows poly-logarithmically with $n$: $\regret{n} \le \widetilde{\mathcal{O}}\left((d+K)^2/\Delta_{\ell}\right)$, where $\Delta_{\ell}$ is an appropriately defined \emph{gap} in the linear model.

\section{Analysis}

To prove Theorem \ref{thm:mainregretbound}, we need to show that the regret of $\algname$ is bounded under either underlying model. In Lemma \ref{lem:conditionnotviolatedundersimple} we demonstrate that whenever the rewards are generated under the simple model, Condition \ref{test:complextest} is \emph{not violated} with high probability. This ensures that when the data is generated from the simple model, we have that the Boolean variable ${\texttt{Current Model}} = \text{`Simple'}$ throughout the run of the algorithm. Thus, the regret is automatically equal to the regret incurred by the $\UCB$ algorithm, which is meant for simple model instances.

On the other hand, when the data is generated according to the complex model, we demonstrate (in Lemma~\ref{lem:complexmodeltestnotviolated}) that the regret remains appropriately bounded if Condition \ref{test:complextest} is \emph{not violated}. 
If the condition gets violated at a certain round, we switch to the estimates of the complex model, that is, $j_t$. 
This corresponds to a variant of the algorithm $\OFUL$, which is meant for complex instances.
Thus, the regret remains bounded in the subsequent rounds under this event as well (formally proved in Lemma~\ref{lem:regretboundafterfailure}). 

We define below several functions which will be used throughout the proof. These arise naturally by applying the concentration inequalities on terms that appear while controlling the regret.
\begin{subequations}
\begin{align}\label{def:taumin}
&\tmin :=\left(\frac{16}{\loweig^2} + \frac{8}{3\loweig}\right)\log\left(\frac{2dn}{\delta'}\right).\\
&    \updela:= \frac{10}{3}\left(2 + \sigma\sqrt{1+2\log\left(\frac{2Kn}{\delta'}\right) }\right)\left[\log\left( \frac{2dn}{\delta'}\right)+ \sqrt{t\log\left( \frac{2dn}{\delta'}\right)+\log^2\left( \frac{2dn}{\delta'}\right)}\right]. \label{def:up}\\
& \mone : = \sqrt{2\sigma^2\left(\frac{d}{2}\log\left(1+\frac{t}{d}\right)+\log\left(\frac{1}{\delta'}\right)\right)} +1 \label{def:mone}.\\
& \kenv:=\begin{cases} \mone + \Upsilon_{\delta'}(t,n), & \text{if } K< t \le K+\tmin,\\
    \frac{\mone}{\sqrt{1+\loweig\cdot (t-K)/2}}+\frac{\Upsilon_{\delta'}(t,n)}{1+\loweig\cdot (t-K)/2} ,&\text{if }  K+\tmin<t.
    \end{cases} \label{def:kenv}\\
& \mathcal{Q}_{\delta'}(t,n) : = 16\sqrt{\log(K)\rho_{max}} \left[\sqrt{\left(\sum_{s=K+1}^{t} (\mathcal{K}_{\delta'}(s-1,n))^2 \right) \log\left(\frac{1}{\delta'}\right)}+ \sum_{s=K+1}^{t}\mathcal{K}_{\delta'}(s-1,n) \right] \nonumber\\ &    \qquad \qquad    \qquad  \qquad \qquad \qquad    \qquad  \qquad\qquad \qquad    \qquad    \qquad  \qquad +3\log\left(\frac{1}{\delta'}\right). \label{def:qenv}\\
     & \wenvone : = 2\mathcal{Q}_{\delta'}(t,n)+\sigma\sqrt{\frac{1+t}{2}\log\left(\frac{1}{\delta'}\right)}  +\left[(2\sigma+3)\sqrt{1+ 2\log\left(\frac{Kt^{1/2}}{\delta'}\right)}\right]\sqrt{Kt}. \label{def:wenv}
\end{align}
\end{subequations}

Given the definitions above, it is easy to verify that $\mathcal{W}_{\delta/n}(n,n) = \widetilde{\mathcal{O}}\left(\sigma(\sqrt{\frac{\higheig}{\loweig}}\sqrt{d}+\sqrt{K})\sqrt{n}\right)$ for 
any \textit{fixed} $\delta \in (0,1)$.

Additionally, we define several statistical events that will be useful in proofs of the lemmas that follow.
\begin{subequations}
\begin{align}
    \mathcal{E}_1 &:=\left\{\left\lvert\sum_{s=K+1}^{t-1}\eta_{i_s,s}\right\rvert \le \sigma\sqrt{\frac{t}{2}\log\left(\frac{1}{\delta'}\right)}, \forall t\in \{K+2,\ldots,n\}\right\}, \label{def:e1}\\
    \mathcal{E}_2 &:= 
   \text{Under the simple model:}  \label{def:e2}\\& \left\{  \lvert \mu_i -\averagerew \rvert \le \sigma\sqrt{\frac{1+T_{i}(t)}{T_{i}^2(t)}\left(1+ 2\log\left(\frac{K(1+T_i(t))^{1/2}}{\delta'}\right)\right)},\forall i\in [K], \text{ and } \forall t\in [n]\right\},\nonumber\\
   &\text{Under the complex model:}  \nonumber\\& \left\{  \lvert \mu_i -\averagerew \rvert \le (\sigma+1)\sqrt{\frac{1+T_{i}(t)}{T_{i}^2(t)}\left(1+ 2\log\left(\frac{K(1+T_i(t))^{1/2}}{\delta'}\right)\right)},\forall i\in [K], \text{ and } \forall t\in [n]\right\},\nonumber\\
    \mathcal{E}_3 &: = \left\{\max\left\{\sum_{s=K+1}^{t} \langle \alpha_{j_s,s},\tilde{\theta}_s - \theta^* \rangle,\sum_{s=K+1}^{t} \langle \alpha_{\kappa_s,s}, \theta^*-\tilde{\theta}_s  \rangle  \right\} \le \mathcal{Q}_{\delta'}(t,n) , \forall t\in\{K+1,\ldots,n\}\right\}.\label{def:e3}
\end{align}
\end{subequations}
Event $\mathcal{E}_1$ represents control on the fluctuations due to noise: applying Theorem \ref{thm:selfnormalized} in the one-dimensional case with $V=1$ and $Y_s = 1$, we get $\mathbb{P}(\mathcal{E}_1^c) \le \delta'$ for all $t\ge 0$. 
Event $\mathcal{E}_2$ represents control on the fluctuations of the empirical estimate of the biases $[\mu_1, \ldots, \mu_K]$ around their true values: by Lemma \ref{lem:csimplebound} we have $\mathbb{P}(\mathcal{E}_2^c) \le \delta'$ in the simple model and by Lemma~\ref{lem:updatedbiasbound} we have $\mathbb{P}(\mathcal{E}_2^c) \le 2\delta'$ in the complex model. 
Finally, event $\mathcal{E}_3$ represents control on the fluctuations of the inner product of the context chosen by the complex model with the difference between $\tilde{\theta}_s$ and $\theta^*$: by Lemma \ref{l:concentrationofalphatimestheta}, we have $\mathbb{P}(\mathcal{E}_3^c) \le 14\delta' n$.
We define the desired event $\mathcal{E} := \mathcal{E}_1\cap\mathcal{E}_2\cap\mathcal{E}_3$ as the intersection of these three events. 
The union bound gives us $\mathbb{P}(\mathcal{E}^c)\le 17\delta' n$. 
For the rest of the proof, we condition on the event $\mathcal{E}$.
Later, we will designate $\delta := \delta' n$, which will give us $\mathbb{P}(\mathcal{E}^c)\le 17\delta$.

\subsection{Regret under the simple model}
The following lemma establishes that under the simple model, Condition \ref{test:complextest} is not violated with high probability.
\begin{lemma} \label{lem:conditionnotviolatedundersimple}
Assume that rewards are generated under the simple model.
Then, with probability at least $1-17\delta' n$, we have
\begin{align*}
   \sum_{s=K+1}^{t-1} \left[\muit{j_s}{s-1} + \langle \alpha_{j_s,s},\tilde{\theta}_s\rangle\right] - \sum_{s=K+1}^{t-1} g_{i_s,s} < \wenv, \qquad \forall t\in \{K+2,\ldots,n\}.
\end{align*}
\end{lemma}
\begin{Proof} Under the simple model, We have the model for the rewards is $g_{i,t} =\mustar{i} + \eta_{i,t}$.
Therefore, we have
\begin{align*}
  &\sum_{s=K+1}^{t-1} \left[\muit{j_s}{s-1} + \langle \alpha_{j_s,s},\tilde{\theta}_s\rangle\right]- \sum_{s=K+1}^{t-1} g_{i_s,s}\\ &\qquad  \quad  = \sum_{s=K+1}^{t-1} \left[\muit{j_s}{s-1} + \langle \alpha_{j_s,s},\tilde{\theta}_s\rangle\right] - \sum_{s=K+1}^{t-1} \mustar{i_s}- \sum_{s=K+1}^{t-1} \eta_{i_s,s}\\
  & \qquad \quad  = \sum_{s=K+1}^{t-1} -\eta_{i_s,s} + \sum_{s=K+1}^{t-1}\left[\muit{i_s}{s-1} - \mustar{i_s}\right] + \sum_{s=K+1}^{t-1} \left[\muit{j_s}{s-1} - \muit{i_s}{s-1}\right]+ \sum_{s=K+1}^{t-1}  \langle \alpha_{j_s,s},\tilde{\theta}_s\rangle\\
  &\qquad \quad  = \underbrace{ \sum_{s=K+1}^{t-1} -\eta_{i_s,s}}_{=: \Gamma_{no}} + \underbrace{\sum_{s=K+1}^{t-1}\left[\muit{i_s}{s-1} - \mustar{i_s}\right]}_{=: \Gamma_{sim1}} + \underbrace{\sum_{s=K+1}^{t-1}\left[\muit{j_s}{s-1} - \muit{i_s}{s-1}\right]}_{=: \Gamma_{sim2}} + \underbrace{\sum_{s=K+1}^{t-1}  \langle \alpha_{j_s,s},\tilde{\theta}_s\rangle}_{=:\Gamma_{lin}}\\
    &\qquad \quad   = \Gamma_{no} + \Gamma_{sim1}+\Gamma_{sim2} + \Gamma_{lin}.
\end{align*}
Notice that the difference neatly decomposes into four terms, each of which we interpret below. 
The first term $\Gamma_{no}$ is purely a sum of the noise in the problem that concentrates under the event $\mathcal{E}_1$. 
The second term $\Gamma_{sim1}$ corresponds to the difference between the true mean reward $\mustar{i_s}$ and simple estimate of the mean reward $\muit{i_s}{s-1}$, which is controlled under the event $\mathcal{E}_2$. 
The third term $\Gamma_{sim2}$ is the difference between the mean rewards prescribed by the simple estimate and complex estimate $\muit{i_s}{s-1}$ and $\muit{j_s}{s-1}$ respectively. 
Finally, the last term $\Gamma_{lin}$ is only a function the estimated linear predictor (and since the true predictor is $\thetastar = 0$, this term is controlled by even $\mathcal{E}_3$). 

\textbf{Step (i)} \emph{(Bound on $\Gamma_{no}$)}: 
Under the event $\mathcal{E}_1$, we have
\begin{align*}
    \Gamma_{no}   \le \sigma\sqrt{\frac{t}{2}\log\left(\frac{1}{\delta'}\right)}.
\end{align*}

\textbf{Step (ii)} \emph{(Bound on $\Gamma_{sim1}$)}: 
 By the definition of $\muit{i}{s-1}$ we have,
\begin{align*}
    \Gamma_{sim1} & = \sum_{s=K+1}^{t-1} \muit{i_s}{s-1} -\mustar{i_s}\\
    & \overset{(i)}{\le} 2 \sigma\sum_{s=K+1}^{t-1} \sqrt{\frac{1+T_{i_s}(s-1)}{T_{i_s}^2(s-1)}\left(1+ 2\log\left(\frac{K(1+T_{i_s}(s-1))^{1/2}}{\delta'}\right)\right)}\\
    &\le 2 \sigma\sum_{s=K+1}^{t-1} \sqrt{\frac{1+T_{i_s}(s-1)}{T_{i_s}^2(s-1)}\left(1+ 2\log\left(\frac{K(t-1)^{1/2}}{\delta'}\right)\right)},
\end{align*}
where $(i)$ follows under the event $\mathcal{E}_2$. Analyzing further we have that
\begin{align*}
    \Gamma_{sim1}&\le \left[2\sigma\sqrt{\left(1+ 2\log\left(\frac{K(t-1)^{1/2}}{\delta'}\right)\right)}\right] \sum_{i=1}^K \sum_{r=1}^{T_i(t-2)}\sqrt{\frac{1+r}{r^2}} \\
    & \le \left[2\sigma\sqrt{\left(1+ 2\log\left(\frac{K(t-1)^{1/2}}{\delta'}\right)\right)}\right] \sum_{i=1}^K \sum_{r=1}^{T_i(t-2)}2\sqrt{\frac{1}{r}}\\
    &\overset{(i)}{\le} \left[2\sigma\sqrt{\left(1+ 2\log\left(\frac{K(t-1)^{1/2}}{\delta'}\right)\right)}\right] \sum_{i=1}^K \sqrt{T_i(t-2)}\\
    &\overset{(ii)}{\le }\left[2\sigma\sqrt{\left(1+ 2\log\left(\frac{K(t-1)^{1/2}}{\delta'}\right)\right)}\right]\sqrt{K(t-1)}, \label{eq:analyzingthebias} \numberthis
\end{align*}
where $(i)$ follows as \begin{align*}
    2\sum_{r=1}^{T_i(t-2)}\sqrt{\frac{1}{r}}\le 2\int_{0}^{T_{i}(t-2)}\sqrt{\frac{1}{r}} \le \sqrt{T_i(t-2)},
\end{align*} and $(ii)$ follows by Jensen's inequality and the fact that $\sum_{i=1}^K T_i(t-2) = t-2 < t-1$.

\textbf{\textbf{Step (iii)}} \emph{(Bound on $\Gamma_{sim2}$)}: Eq.~\eqref{eq:simpleestimate}, which shows the optimality of arm $i_s$, tells us that $\muit{i_s}{s-1} \ge \muit{j_s}{s-1}$ for all $s$. Therefore $\Gamma_{sim2}\le 0$.

\textbf{Step (iv)} \emph{(Bound on $\Gamma_{lin}$)}: By the definition of event $\mathcal{E}_3$ (since $\theta^* = 0$ under the simple model)
\begin{align*}
    \Gamma_{lin} = \sum_{s=K+1}^{t-1} \langle \alpha_{j_s,s},\tilde{\theta}_s\rangle \le \mathcal{Q}_{\delta'}(t-1,n)
\end{align*} 
where $\mathcal{Q}_{\delta'}(t-1,n)$ is defined in Eq.~\eqref{def:qenv}. 

Combining the bounds on $\Gamma_{no}, \Gamma_{sim1},\Gamma_{sim2}$ and $\Gamma_{lin}$ and by the definition of $\wenv$, we have
\begin{align*}
     \sum_{s=K+1}^{t-1} \left[\muit{j_s}{s-1} + \langle \alpha_{j_s,s},\tilde{\theta}_s\rangle\right] - \sum_{s=K+1}^{t-1} g_{i_s,s} &\le   \wenv,
\end{align*}
which completes the proof.
\end{Proof}
\begin{Proof}[of Part (a) of Theorem \ref{thm:mainregretbound}]
We have established above that Condition \ref{test:complextest} is \emph{not violated} with probability at least $1-17 \delta' n$ under the simple model by the lemma above. Conditioned on this event, $\algname$ plays according to the simple model estimate, $i_t$, for all rounds. Invoking Theorem~7 in \citep{abbasi2011improved} gives us that with probability at least $1-\delta'$, $\psued{n}\le \sum_{i:\Delta_i >0} 3\Delta_i + (16/\Delta_i)\log(2K/\Delta_i \delta')$. Applying the union bound over these two events gives this regret bound with probability at least $1-18 \delta' n$.
Finally, setting $\delta := \delta'n$ and plugging it in to the above inequality gives us the statement of Equation~\eqref{eq:thmsm} with probability at least $1 - 18\delta$.
\end{Proof}

\subsection{Regret under the complex model}

The bound on the regret under the complex model follows by establishing two facts. First, when Condition~\ref{test:complextest} is not violated, we demonstrate in Lemma~\ref{lem:complexmodeltestnotviolated} that the regret is appropriately bounded.
Second, if the condition does get violated, say at round $\tau_*$, our algorithm $\algname$ chooses arms according to the complex model estimates `$j_t$' for $t \in [\tau_*,\ldots, n]$.
In Lemma~\ref{lem:regretboundafterfailure}, we show that the regret remains bounded in this case as well.

We start with the first case by stating and proving Lemma~\ref{lem:complexmodeltestnotviolated}.

\begin{lemma} \label{lem:complexmodeltestnotviolated} For all $t \in \{K+1, \ldots,  n\}$.
Let Condition \ref{test:complextest} not be violated up until round $t+1$, that is,
\begin{align*}
\sum_{s=K+1}^{t} \left\{\muit{j_{s}}{s-1} + \langle \alpha_{j_s,s},\tilde{\theta}_s\rangle -  g_{i_{s},s}\right\} \le \wenvone .
\end{align*}
Then, we have
\begin{align*}
    \regret{t} \le 4K+ 2\wenvone 
\end{align*}
with probability at least $1 - 17 \delta' n$.
\end{lemma}
\begin{Proof} 
Since we have already conditioned on the event $\mathcal{E}$, we can assume that events $\mathcal{E}_1$, $\mathcal{E}_2$ and $\mathcal{E}_3$ hold.
Note that if Condition \ref{test:complextest} is not violated up to round t then we have that $A_s = i_s$ for all $s \leq t$. Using the definition of $\regret{t}$, we get
\begin{align*}
    \regret{t} & = \sum_{s=1}^t \left[\mu_{\kappa_s}+\langle \theta^*,\alpha_{\kappa_s,s}\rangle - \mu_{i_s}-\langle \theta^*,\alpha_{i_s,s}\rangle \right]\\
    &\le 4K + \sum_{s=K+1}^t \left[\mu_{\kappa_s}+\langle \theta^*,\alpha_{\kappa_s,s}\rangle - \mu_{i_s}-\langle \theta^*,\alpha_{i_s,s}\rangle \right]\\
    & = 4K+\sum_{s=K+1}^t \left( \mu_{\kappa_s}+\langle \theta^*,\alpha_{\kappa_s,s}\rangle - g_{i_s,s}\right) + \sum_{s=K+1}^t \left(g_{i_s,s} - \mu_{i_s}-\langle \theta^*,\alpha_{i_s,s}\rangle   \right)\\
    & = 4K+\sum_{s=K+1}^t \left(\mu_{\kappa_s}+\langle \theta^*,\alpha_{\kappa_s,s}\rangle  - \muit{j_s}{s-1} -\langle \tilde{\theta}_{s},\alpha_{j_s,s}\rangle \right) \\&\qquad \qquad \qquad \qquad \qquad \qquad \qquad  + \underbrace{\sum_{s=K+1}^t \left(\muit{j_s}{s-1} +\langle \tilde{\theta}_{s},\alpha_{j_s,s}\rangle - g_{i_s,s} \right)}_{\le \wenvone}+  \sum_{s=K+1}^t\eta_{i_s,s} \\
    & \le 4K + \wenvone+\underbrace{\sum_{s=K+1}^t \left(\mu_{\kappa_s}+\langle \theta^*,\alpha_{\kappa_s,s}\rangle  - \muit{j_s}{s-1} -\langle \tilde{\theta}_{s},\alpha_{j_s,s}\rangle \right)}_{=:\Gamma_{lin}} + \underbrace{\sum_{s=K+1}^t\eta_{i_s,s}}_{=:\Gamma_{no}},
\end{align*}
where $4K$ is the maximum possible regret incurred in the first $K$ rounds under the complex model. By the definition of $\mathcal{E}_1$, we get $\Gamma_{no} \le \sigma \sqrt{((1+t)/2)\log(1/\delta')}$. Next, let us control $\Gamma_{lin}$.
We have
\begin{align*}
    \Gamma_{lin} &= \sum_{s=K+1}^t \left(\mu_{\kappa_s}+\langle \theta^*,\alpha_{\kappa_s,s}\rangle  - \muit{j_s}{s-1} -\langle \tilde{\theta}_{s},\alpha_{j_s,s}\rangle \right) \\
    & = \sum_{s=K+1}^t \left(\mu_{\kappa_s}+\langle \theta^*,\alpha_{\kappa_s,s}\rangle  - \muit{\kappa_s}{s-1} -\langle \tilde{\theta}_{s},\alpha_{\kappa_s,s}\rangle\right) \\ & \qquad \qquad \qquad \qquad \qquad \qquad \qquad + \underbrace{\sum_{s=K+1}^t \left(\muit{\kappa_s}{s-1} +\langle \tilde{\theta}_{s},\alpha_{\kappa_s,s}\rangle  - \muit{j_s}{s-1} -\langle \tilde{\theta}_{s},\alpha_{j_s,s}\rangle\right)}_{\le 0},
\end{align*}
where the non-positivity of the second term is because of the optimality of arm $j_s$ as expressed in Eq.~\eqref{eq:complexestimate}. Hence, we have
\begin{align*}
    \Gamma_{lin} &\le \sum_{s=K+1}^t \left(\mu_{\kappa_s}+\langle \theta^*,\alpha_{\kappa_s,s}\rangle  - \muit{\kappa_s}{s-1} -\langle \tilde{\theta}_{s},\alpha_{\kappa_s,s}\rangle\right)\\
    &= \sum_{s=K+1}^t \mu_{\kappa_s} - \muit{\kappa_s}{s-1} + \sum_{s=K+1}^t \langle \alpha_{\kappa_s,s},\theta^* - \tilde{\theta}_s\rangle.
\end{align*}
Under the event $\mathcal{E}_2$ in the complex model we have 
\begin{align*}
    \mu_{\kappa_s} - \muit{\kappa_s}{s-1} & \le  \sum_{s=K+1}^{t-1} \sqrt{\frac{1+T_{\kappa_s}(s-1)}{T_{\kappa_s}^2(s-1)}\left(1+ 2\log\left(\frac{K(1+T_{\kappa_s}(s-1))^{1/2}}{\delta'}\right)\right)}\\
    & \le \left[\sqrt{\left(1+ 2\log\left(\frac{K(t-1)^{1/2}}{\delta'}\right)\right)}\right]\sqrt{K(t-1)},
\end{align*}
where the second inequality follows by mirroring the logic used to arrive at inequality~\eqref{eq:analyzingthebias} above. Also, by the definition of $\tilde{\theta}_s$ and under event $\mathcal{E}_3$, we have
\begin{align*}
    \sum_{s=K+1}^t \langle \alpha_{\kappa_s,s},\theta^* - \tilde{\theta}_s\rangle \le \mathcal{Q}_{\delta'}(t,n).
\end{align*}
Therefore, we have that 
\begin{align*}
    \Gamma_{lin} \le  \left[\sqrt{\left(1+ 2\log\left(\frac{K(t-1)^{1/2}}{\delta'}\right)\right)}\right]\sqrt{K(t-1)}+ \mathcal{Q}_{\delta'}(t,n).
\end{align*}
Combining these bounds, we get
\begin{align*}
    \regret{t} &\le 4K+ \wenvone + \sigma \sqrt{\frac{1+t}{2}\log \left(\frac{1}{\delta'}\right)}\\ & \qquad \qquad +\left[\sqrt{\left(1+ 2\log\left(\frac{K(t-1)^{1/2}}{\delta'}\right)\right)}\right]\sqrt{K(t-1)}+\mathcal{Q}_{\delta'}(t,n)\\&\le 4K+2\wenvone
\end{align*}
under the assumption that event $\mathcal{E}$ holds.
Since we already showed that $\mathbb{P}(\mathcal{E})\ge 1-17\delta' n$, our proof is complete.
\end{Proof}

Now, we move on to the second case.
The next lemma shows that if Condition \ref{test:complextest} was violated at round $\tau_*$ (which is, in general, a random variable), then playing the complex model estimates $j_s$ for all $s \geq \tau_*$ keeps the regret bounded in subsequent rounds.
\begin{lemma}\label{lem:regretboundafterfailure}If Condition \ref{test:complextest} is violated at round $\tau_*$ that is,
$$\sum_{s=K+1}^{\tstar-1} \left\{\muit{j_{s}}{s-1} + \langle \alpha_{j_s,s},\tilde{\theta}_s\rangle -  g_{i_{s},s}\right\} > \wenvtau.$$
Then with probability at least $1-16\delta' n$ we have
\begin{align*}
    R^c_{\tau_{*}:n} : = \sum_{s=\tau_*}^t \left[\mu_{\kappa_s}+\langle \theta^*,\alpha_{\kappa_s,s}\rangle - \mu_{A_s}-\langle \theta^*,\alpha_{A_s,s}\rangle \right] \le 2\wenvn.
\end{align*}
\end{lemma}
\begin{Proof} 
For this proof, we only need events $\mathcal{E}_2$ and $\mathcal{E}_3$ to simultaneously hold.
We define the event $\mathcal{E}' := \mathcal{E}_2\cap\mathcal{E}_3$.
Again, by the union bound we have $\mathbb{P}(\mathcal{E}^c)\le 16\delta' n$. 
For the rest of this proof we assume the event $\mathcal{E}'$.

If Condition~\ref{test:complextest} is violated at round $\tau^*$, then we have $A_s = j_s$ for all rounds $s\ge\tau_*$. 
Thus,
\begin{align*}
     R^c_{\tau_{*}:n} & = \sum_{s=\tau_*}^n \left[\mu_{\kappa_s}+\langle \theta^*,\alpha_{\kappa_s,s}\rangle - \mu_{j_s}-\langle \theta^*,\alpha_{j_s,s}\rangle \right] \\
     & = \sum_{s=\tau_*}^n \left[\mu_{\kappa_s}+\langle \theta^*,\alpha_{\kappa_s,s}\rangle - \muit{j_s}{s-1}-\langle \tilde{\theta}_s,\alpha_{j_s,s}\rangle \right]\\ & \qquad \qquad \qquad  + \sum_{s=\tau_*}^n \left[\muit{j_s}{s-1}+\langle \tilde{\theta}_s,\alpha_{j_s,s}\rangle - \mu_{j_s}-\langle \theta^*,\alpha_{j_s,s}\rangle \right] \\
     & = \sum_{s=\tau_*}^n \left[\mu_{\kappa_s}+\langle \theta^*,\alpha_{\kappa_s,s}\rangle - \muit{\kappa_s}{s-1}-\langle \tilde{\theta}_s,\alpha_{\kappa_s,s}\rangle \right]\\
     &\qquad \qquad \qquad  +\underbrace{\sum_{s=\tau_*}^n \left[\muit{\kappa_s}{s-1}+\langle \tilde{\theta}_s,\alpha_{\kappa_s,s}\rangle - \muit{j_s}{s-1}-\langle \tilde{\theta}_s,\alpha_{j_s,s}\rangle \right]}_{\le 0}\\&\qquad \qquad \qquad \qquad \qquad \qquad  + \sum_{s=\tau_*}^n \left[\muit{j_s}{s-1}+\langle \tilde{\theta}_s,\alpha_{j_s,s}\rangle - \mu_{j_s}-\langle \theta^*,\alpha_{j_s,s}\rangle \right],
\end{align*}
where the second term is non-positive by the optimality of arm $j_s$ as expressed in Eq.~\eqref{eq:complexestimate}.  Therefore, we get
\begin{align*}
    R^c_{\tau_{*}:n} & \le \underbrace{\sum_{s=\tau_*}^n \left[\langle \alpha_{\kappa_s,s} , \theta^* - \tilde{\theta}_s\rangle + \langle \alpha_{j_s,s}, \tilde{\theta}_s- \theta^*  \rangle\right]}_{=: \Gamma_{lin}}+ \underbrace{\sum_{s=\tau_*}^n \muit{j_s}{s-1} - \mu_{j_s}+\sum_{s=\tau^*}^n \mu_{\kappa_s}-\tilde{\mu}_{\kappa_s,s-1}}_{\Gamma_{bias}},
\end{align*}
First we control $\Gamma_{lin}$.
Under the event $\mathcal{E}_3$ we have 
\begin{align*}
    \Gamma_{lin} =\sum_{s=\tau_*}^n \left[\langle \alpha_{\kappa_s,s} , \theta^* - \tilde{\theta}_s\rangle + \langle \alpha_{j_s,s}, \tilde{\theta}_s- \theta^*  \rangle\right] \le 2 \mathcal{Q}_{\delta'}(n,n).
\end{align*}

Next, we control the term $\Gamma_{bias}$.
By the definition of $\muit{j_s}{s-1}$ and under event $\mathcal{E}_2$ in the complex model we have the first term
\begin{align*}
    \sum_{s=\tau_*}^n \muit{j_s}{s-1} - \mu_{j_s}
    & \le 2 (\sigma+1)\sum_{s=\tau_*}^n \sqrt{\frac{1+T_{j_s}(s-1)}{T_{j_s}^2(s-1)}\left(1+ 2\log\left(\frac{K(1+T_{j_s}(s-1))^{1/2}}{\delta'}\right)\right)}\\
    &\le  2 (\sigma+1)\sum_{s=\tau_*}^n \sqrt{\frac{1+T_{j_s}(s-1)}{T_{j_s}^2(s-1)}\left(1+ 2\log\left(\frac{Kn^{1/2}}{\delta'}\right)\right)}\\
    & \le \left[ 2 (\sigma+1)\sqrt{1+ 2\log\left(\frac{K(n^{1/2}}{\delta'}\right)}\right] \sum_{i=1}^K \sum_{r=1}^{T_i(n-1)}\sqrt{\frac{1+r}{r^2}} \\
    & \le \left[ 2 (\sigma+1)\sqrt{1+ 2\log\left(\frac{Kn^{1/2}}{\delta'}\right)}\right] \sum_{i=1}^K \sum_{r=1}^{T_i(n-1)}2\sqrt{\frac{1}{r}}\\
    &\le \left[ 2 (\sigma+1)\sqrt{1+ 2\log\left(\frac{Kn^{1/2}}{\delta'}\right)}\right] \sum_{i=1}^K \sqrt{T_i(n-1)}\\
    &\overset{(i)}{\le }\left[ 2 (\sigma+1)\sqrt{1+ 2\log\left(\frac{Kn^{1/2}}{\delta'}\right)}\right]\sqrt{Kn},
\end{align*}
where $(i)$ follows by Jensen's inequality and the fact that $\sum_{i=1}^K T_i(n-1) = n-1 < n$.
The rest of the inequalities can be verified by some simple algebra.
By using similar logic the second term in $\Gamma_{bias}$ is bounded by
\begin{align*}
    \sum_{s = \tau^* }^n \mu_{\kappa_s}- \muit{\kappa_s}{s-1} \le \sqrt{1+ 2\log\left(\frac{Kn^{1/2}}{\delta'}\right)}\sqrt{Kn}.
\end{align*}
Combining the bounds on the respective terms, we get
\begin{align*}
    R^c_{\tau_{*}:n} &\le 2\mathcal{Q}_{\delta'}(n,n) + \left[ (2 \sigma+3)\sqrt{\left(1+ 2\log\left(\frac{Kn^{1/2}}{\delta'}\right)\right)}\right]\sqrt{Kn} \le 2\wenvn,
\end{align*}
which completes the proof.
\end{Proof}

Armed with these two lemmas, we are now ready to complete the proof of Part (b) of Theorem \ref{thm:mainregretbound}, and bound the regret under the complex model.
\begin{Proof}[of Part (b) of Theorem \ref{thm:mainregretbound}]
One out the two disjoint events are possible under the complex model.

\textbf{Case 1:} In the first event Condition~\ref{test:complextest} is never violated throughout the run of the algorithm. Then by Lemma \ref{lem:complexmodeltestnotviolated} we have
\begin{align*}
    \regret{n} \le 4K+2\wenvn
\end{align*}
with probability at least $1 - 17\delta' n$.

\textbf{Case 2:} The other event is when Condition \ref{test:complextest} is violated in round $\tau_{*}<n$ (for some random time $\tau_{*}$). We know by Lemma~\ref{lem:complexmodeltestnotviolated} that
\begin{align*}
\regret{\tau_*-2} \le 4K+2\wenvn
\end{align*}
with probability at least $1 - 17\delta' n$.
Also, by Lemma \ref{lem:regretboundafterfailure}, we have
\begin{align*}
R^c_{\tau_{*}:n} := \sum_{s=\tau_*}^t \left[\mu_{\kappa_s}+\langle \theta^*,\alpha_{\kappa_s,s}\rangle - \mu_{A_s}-\langle \theta^*,\alpha_{A_s,s}\rangle \right] &\le 2\wenvn
\end{align*}
with probability at least $1 - 16\delta' n$.
We can decompose the cumulative regret up to round $n$ as follows:
\begin{align*}
    \regret{n} \le \regret{\tau_{*}-1} + R^c_{\tau_{*}:n} + 4,
\end{align*}
where $R^c_{\tau_{*}:n}$ denotes the regret of the algorithm starting from round $\tau^*$ up to round $n$ and the $4$ appears as it is the maximum regret that could be incurred in round $\tstar$ by the algorithm under the complex model. By taking a union bound and using the decomposition of the regret above, we get $\regret{n}\le 4(K+1)+4\wenvn,$
with probability at least $1-33\delta' n$.
Setting $\delta := \delta' n$ then gives us precisely the statement of Equation~\eqref{eq:thmcm} with probability at least $1 - 33\delta$.
\end{Proof}

\section{Experiments}
\begin{figure}[t]
\centering
  \includegraphics[width=0.95\linewidth]{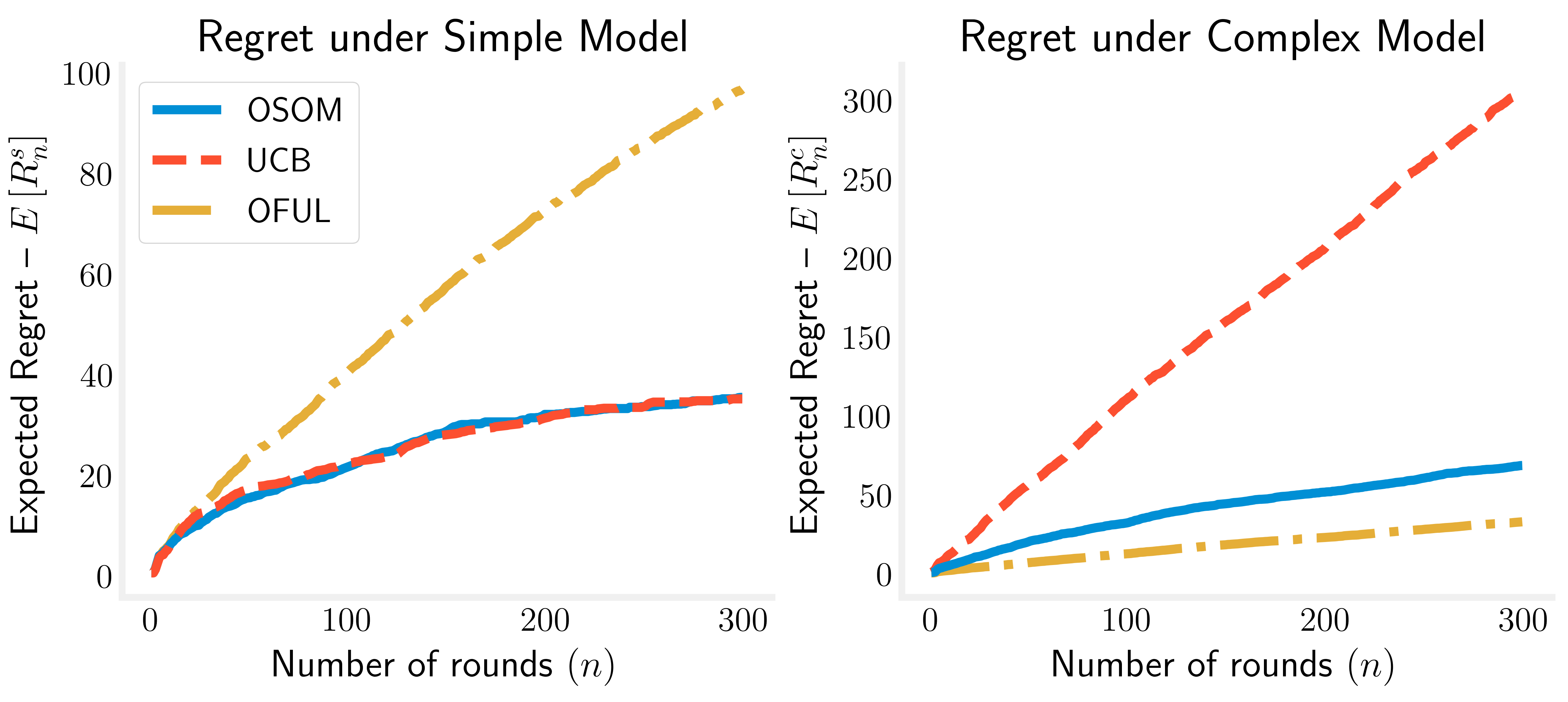}
  \label{fig:simple}
\caption{Experiments on synthetic data with $K=5$, $d=50$ and number of rounds was $n=300$. The three algorithms that we ran were $\algname$, $\UCB$ and $\OFUL$.
}
\label{fig:test}
\end{figure}
To experimentally corroborate our claims, we ran our model-selecting algorithm, $\algname$, on both simple and complex instances. 
We compared its performance to that of $\UCB$ (which is optimal up to logarithmic factors under the simple model) and $\OFUL$ (which is minimax optimal under the complex model). The data was generated synthetically with the number of arms $K= 5$, and the dimension of $\theta^*$, $d = 50$.

The mean rewards of the arms $\mu_i$ 
were drawn independently from a uniform distribution on $(-1,1)$, and the context vectors $\alpha_{i,t}$ were drawn independently from the uniform distribution over the sphere. The noise $\eta_{i,t}$ 
was drawn from a 1-dimensional Gaussian with unit variance. Under the simple model $\theta^* = 0$, while under the complex model $\theta^*$ was also drawn from the uniform distribution over the unit sphere in $d$-dimensions. In both the experiments we average over $50$ runs over $n=300$ rounds to estimate the expected regret incurred. The realizations of the problem were drawn independently for each run of each algorithm. For both $\OFUL$ and $\algname$ we used the empirical covariance matrix to build the upper confidence ellipsoid.

When data is generated according to the simple model ($\theta^* = 0$), we see that $\algname$ and $\UCB$ suffer regret that is sub-linear, and is significantly lower than the regret suffered by $\OFUL$ whose regret is also sub-linear but pays for the additional variance of estimating a more complex model. When the data is generated from the complex model ($\lv \theta^* \rv_2 = 1$) the regret suffered by $\UCB$ is \emph{linear}, as it does not identify and estimate the linear structure of the mean rewards. Here, the regret suffered by both $\OFUL$ and $\algname$ is sub-linear and almost identical.
\section{Discussion}
We were able to successfully obtain minimax-optimal rates in both regimes under suitable stochastic conditions on the contextual information. This is a natural step to understanding data-dependent model selection for contextual bandits. A number of exciting directions remain open. 
\begin{itemize}[leftmargin=*]
\item We crucially relied on the linear structure of the rewards to obtain our regret bounds. It is conceivable that this \emph{linearity} is not essential, and that these algorithmic ideas could be generalized to arbitrary nested models.
\item Another interesting direction would be to investigate bounds on overall reward when the data is generated from a richer model that is not from a linear model or a simple bandit model, but can be reasonably approximated by it.
\item Our guarantees here are under a stochastic assumption on both the rewards and the distribution of the contexts. It would be interesting to understand whether these assumptions can be loosened, or if there exist fundamental limitations to model-selecting under bandit feedback in adversarial settings.
\end{itemize}
\subsection*{Acknowledgements}
The authors would like to thank Kush Bhatia, Akshay Krishnamurthy and Anant Sahai for helpful initial discussions, and Weihao Kong and Avishek Ghosh for comments that led to improvements in sufficient conditions for the results in this paper. We gratefully acknowledge the support of the NSF through grants AST-1444078, IIS-1619362 and ECCS-1343398, and to ML4Wireless center member companies. This work was done in part while the authors were visiting the Simons Institute for the Theory of Computing.
\newpage
\appendix
\flushleft{\textbf{\LARGE{Appendix}}}
\section{Omitted Proof Details}\label{app:omittedproofs}
We recall Lemma \ref{prop:thetaconfidence}, which is an error bound on the ridge regression estimate $\That$, and present a proof below.
\thetaset*
\begin{Proof}  To unclutter notation, let $\Aboldtn = \Aboldt, \Gboldtn = \Gboldt$. Further, define $\etabold = [\eta_{A_{K+1},K+1},\ldots,\eta_{A_t,t}]^{\top}$, $\mubold=[\mu_{A_{K+1}},\ldots,\mu_{A_t}]^{\top}$ and $\mutbold=[\muit{A_{K+1}}{K},\ldots,\muit{A_t}{t-1}]^{\top}$. By the definition of $\That$, we have
\begin{align*}
    \That &= \left(\Aboldtn^{\top}\Aboldtn +  I\right)^{-1} \Aboldtn^{\top} \Gboldtn\\
    &=\left(\Aboldtn^{\top}\Aboldtn +  I\right)^{-1} \Aboldtn^{\top} \left( \Aboldtn \theta^* + (\mubold - \mutbold + \etabold)\right)\\
    &= \theta^* - \left(\Aboldtn^{\top}\Aboldtn +  I\right)^{-1}\theta^* + \left(\Aboldtn^{\top}\Aboldtn +  I\right)^{-1} \Aboldtn^{\top}  \left(\mubold - \mutbold \right) + \left(\Aboldtn^{\top}\Aboldtn +  I\right)^{-1}\Aboldtn^{\top} \etabold.
\end{align*}
Now, let us define $V_t : = \Aboldtn^{\top}\Aboldtn +  I$.
Then, for any vector $w \in \mathbb{R}^d$ (whose choice we will specify shortly), we get
\begin{align*}
    w^{\top}\left(\That - \theta^*\right) & = - w^{\top}V_t^{-1}\theta^* + w^{\top}V_t^{-1} \Aboldtn^{\top}  \left(\mubold - \mutbold \right) + w^{\top}V_t^{-1}\Aboldtn^{\top}\etabold \\
    & = - w^{\top}V_t^{-1/2} V_t^{-1/2}\theta^* + w^{\top}V_t^{-1/2} V_t^{-1/2} \Aboldtn^{\top}  \left(\mubold - \mutbold \right) + w^{\top}V_t^{-1/2} V_t^{-1/2}\Aboldtn^{\top}\etabold .
\end{align*}
By the Cauchy-Schwarz inequality, we have
\begin{align}
    \nonumber\left\lvert w^{\top}\left(\That - \theta^*\right)\right \rvert &\le \lv w \rv_{V_t^{-1}}\left(\lv \Aboldtn^{\top}\etabold\rv_{V_t^{-1}} + \lv \theta^*\rv_{V_t^{-1}}+ \lv\Aboldtn^{\top}\left(\mubold - \mutbold \right)\rv_{V_t^{-1}}\right),\\
    \label{eq:wupperbound}&\le \lv w \rv_{V_t^{-1}}\left(\lv \Aboldtn^{\top}\etabold\rv_{V_t^{-1}} + \lv\Aboldtn^{\top}\left(\mubold - \mutbold \right)\rv_{V_t^{-1}}+ 1\right), 
\end{align}
where the second step follows as $\lv \theta^*\rv_{V_t^{-1}}\le \sqrt{(1/\gamma_{\min}(V_t))}\cdot \lv \theta^*\rv_2 \le 1$. We now define three events $\mathcal{E}_4,\mathcal{E}_5$ and $\mathcal{E}_6$ below:
\begin{align*}
    \mathcal{E}_4 &:=\left\{\lv \Aboldtn^{\top}\etabold\rv_{V_t^{-1}} \le \sqrt{2\sigma^2 \log\left( \frac{\det(V_t)^{1/2}}{\delta'}\right)}, \forall t\in\{K+1,\ldots,n\}\right\},\\
    \mathcal{E}_5 &:= \left\{ N_t:= \left\lv \sum_{s=K+1}^{t} \alpha_{A_s,s}\left(\mu_{A_s}-\muit{A_s}{t-1} \right)\right\rv_2 \le \updela,\forall t\in\{K+1,\ldots,n\}\right\},\\
    \mathcal{E}_6 &: = \left\{\gamma_{\min}(V_t) \ge 1+\loweig(t-K)/2, \forall t\in\{K+\tmin,\ldots,n\}\right\}.
\end{align*}
Define the event $\mathcal{E}'' := \mathcal{E}_4 \cap\mathcal{E}_5 \cap\mathcal{E}_6$. By Theorem \ref{thm:selfnormalized} with $V=I$ we have, $\mathbb{P}(\mathcal{E}_4^c) \le \delta'$, by Lemma \ref{lem:controlmagnitude} we have $\mathbb{P}(\mathcal{E}_5^c) \le \delta'$ and Lemma \ref{lem:eigenvaluelowerbound} tells us that $\mathbb(P)(\mathcal{E}_6^c)\le \delta'$. Therefore by a union bound $\mathbb{P}(\mathcal{E}^c)\le 3\delta'$. For the rest of the proof, we assume the event $\mathcal{E}''$. Hence, we get
\begin{align*}
    \lv \Aboldtn^{\top}\etabold\rv_{V_t^{-1}} &\le \sqrt{2\sigma^2 \log\left( \frac{\det(V_t)^{1/2}}{\delta'}\right)} \overset{(i)}{\le} \sqrt{2\sigma^2\left(\frac{d}{2}\log\left(1+\frac{t}{d}\right)+\log\left(\frac{1}{\delta'}\right)\right)}, \numberthis \label{eq:noiseupperbound}
\end{align*}
where $(i)$ follows by the technical Lemma \ref{lem:detcontrol}. For the other term, we have
\begin{align}
    \left\lv\Aboldtn^{\top}\left(\mubold - \mutbold \right)\right\rv_{V_t^{-1}} &\le \frac{N_t}{\sqrt{\gamma_{\min}(V_t)}}\le \frac{\updela}{\sqrt{\gamma_{\min}(V_t)}}.
\end{align}
Under $\mathcal{E}_6$, we have
\begin{align*}
     \left\lv\Aboldtn^{\top}\left(\mubold - \mutbold \right)\right\rv_{V_t^{-1}} \le \begin{cases} \updela, & \text{if } \tau_{\min}(\delta')\ge t-K>0, \\
      \frac{\updela}{\sqrt{1+\loweig(t-K)/2}}, & \text{if } t-K > \tau_{\min}(\delta'). \numberthis \label{eq:biasupperbound}
     \end{cases}
\end{align*}
Choosing $w = V_t(\That - \theta^*)$ and plugging in the upper bounds established in Eq. \eqref{eq:noiseupperbound} and Eq. \eqref{eq:biasupperbound} into Eq. \eqref{eq:wupperbound}, we get
\begin{align*}
    \lv \That- \theta^*\rv_{V_t} \le\begin{cases} \mone+ \updela , & \text{if } \tmin\ge t-K>0,\\
    \mone+\frac{\updela}{\sqrt{1+\loweig(t-K)/2}} ,& \text{if }  t-K>\tmin.
    \end{cases}
\end{align*}
Recall the definition of $\mone$ in Eq. \eqref{def:mone}. Using the fact that $\lv \That-\theta^* \rv_2 \le (1/\sqrt{\gamma_{\min}(V_t)})\lv \That-\theta^*\rv_{V_t}$ along with the event $\mathcal{E}_6$, we get
\begin{align*}
    \lv \That- \theta^*\rv_{2} &\le\begin{cases} \mone + \updela , & \text{if } \tmin\ge t-K>0,\\
   \frac{\mone}{\sqrt{1+\loweig(t-K)/2}}+\frac{\updela}{1+\loweig(t-K)/2} ,& \text{if } t-K>\tmin,
    \end{cases}\\ &= \kenv,
\end{align*}
where the last equality is by the definition of $\kenv$ in Eq. \eqref{def:kenv}.
\end{Proof}

Now, we establish a couple of concentration inequalities on quantities of interest in the proof of Lemma~\ref{prop:thetaconfidence}: these constitute Lemmas~\ref{lem:eigenvaluelowerbound} and~\ref{lem:controlmagnitude}.

\begin{lemma}\label{lem:eigenvaluelowerbound} Define the matrix $M_t$ as
\begin{align*}
    M_t := I + \sum_{s=1}^{t} \alpha_{A_s,s}\alpha_{A_s,s}^{\top}.
\end{align*}
Then, with probability at least $1-\delta'$, we have
\begin{align*}
\gamma_{\min}(M_t) \ge 1+ \frac{\loweig t}{2}
\end{align*} for all $\tmin \le t \le n$.
\end{lemma}

\begin{Proof}
Note that by the definition of $M_{t}$, we have $\gamma_{\min}(M_{t}) = 1 + \gamma_{\min}\left(\sum_{s=1}^{t} \alpha_{A_s,s}\alpha_{A_s,s}^{\top}\right)$. By the assumption on the distribution of the contexts as specified in Eq.~\eqref{def:assumptioncontext}, we have $\mathbb{E}_{s-1}\left[\alpha_{A_s,s}\alpha_{A_s,s}^{\top}\right] = \Sigma_c \succeq \loweig I$. 
Consider the matrix martingale defined by 
\begin{align*}Z_t := \sum_{s=1}^{t}\left[\alpha_{A_s,s}\alpha_{A_s,s}^{\top} - \Sigma_c\right] \text{ for $t={1,2,\ldots}$ }
\end{align*}
with $Z_0 = 0$ and the corresponding martingale difference sequence $Y_s := Z_{s} - Z_{s-1}$ for $s=\{1,2,\ldots\}$. As $\lv \alpha_{A_s,s}\rv_2 \le 1$ and $\lv \Sigma_c \rv_{op} = \lv \mathbb{E}_{s-1}\left[\alpha_{A_s,s}\alpha_{A_s,s}^{\top}\right]\rv_{op}\le 1$, we have 
 \begin{align*}
 \lv Y_s \rv_{op} = \lv \alpha_{A_s,s}\alpha_{A_s,s}^{\top} - \Sigma_c \rv_{op}\le 2.
 \end{align*}
We also have
 \begin{align*}
     \left\lv \mathbb{E}_{s-1}\left[Y_sY_s^{\top}\right]\right\rv_{op} = \left\lv \mathbb{E}_{s-1}\left[Y_s^{\top}Y_s\right]\right\rv_{op} &=  \left\lv \mathbb{E}_{s-1}\left[\left(\alpha_{A_s,s}\alpha_{A_s,s}^{\top} - \Sigma_c\right)\left(\alpha_{A_s,s}\alpha_{A_s,s}^{\top} - \Sigma_c \right) \right]\right\rv_{op}\\
     & \le \left\lv \mathbb{E}_{s-1}\left[(\alpha_{A_s,s}^{\top}\alpha_{A_s,s}) \alpha_{A_s,s}\alpha_{A_s,s}^{\top} - \Sigma_c^2\right]\right\rv_{op}
     \le 2.
 \end{align*}
 By applying the Matrix Freedman inequality (Theorem \ref{thm:matrixfreedman} in Appendix \ref{sec:tech}) with $R = 2$, $\omega^2 = 2t$ and $u = \loweig t/2$, we get that if $t \ge \left(16/\loweig^2 + 8/(3\loweig)\right)\log\left(2dn/\delta'\right)$, then 
 \begin{align*}
     \mathbb{P}\left\{\left\lv\sum_{s=1}^t \alpha_{A_s,s}\alpha_{A_s,s}^{\top} - t\cdot \Sigma_c\right\rv_{op} \ge \frac{\loweig t}{2}\right\} \le \frac{\delta'}{n}.
 \end{align*}
 This implies that \begin{align*}\gamma_{\min}\left(\sum_{s=1}^t\alpha_{A_s,s}\alpha_{A_s,s}^{\top}\right) \ge \frac{\loweig t}{2}
 \end{align*}
 for a given $t \in \{\tmin, \ldots, n\}$ with probability at least $(1 - \delta'/n)$.
 Taking a union bound over all $t\in \{\tmin,\ldots,n\}$ yields the desired claim with probability at least $1 - \delta'$.
 This completes the proof.
\end{Proof}

\begin{lemma} \label{lem:controlmagnitude}Define the vector 
$ N_t : = \sum_{s=K+1}^t \alpha_{i_s,s}(\mu_{i_s} - \muit{i_s}{s-1})$.
For all $K<t\le n$ we have
$$ \lv N_t \rv_2 \le \updela
$$
with probability at least $1-\delta'$.
\end{lemma}
\begin{Proof}Consider $K<t\le n$. Note that $\muit{i_s}{s-1}$ is a function of $g_{i_1,1},\ldots, g_{i_{s-1},s-1}$ and $i_1,\ldots,i_{s-1}$. Also, the simple model estimate $i_s$ is just a function of $g_{i_1,1},\ldots, g_{i_{s-1},s-1}$ and $A_1,\ldots,A_{s-1}$. Therefore, we have \begin{align*}\mathbb{E}_{s-1}\left[\alpha_{i_s,s}(\mu_{i_s} - \muit{i_s}{s-1})\right] = (\mu_{i_s} - \muit{i_s}{s-1})\mathbb{E}_{s-1}\left[\alpha_{i_s,s}\right]=0
\end{align*}
for all $s \in \{K+1,\ldots,t\}$, as $\alpha_{i_s,s}$ is assumed to drawn from a distribution with zero (conditional) mean. Recall that $\mu_{i_s} \in [-1,1]$. By the definition $\muit{i_s}{s-1}$, we have
 \begin{align*}
\muit{i_s}{s-1} = \underbrace{\sum_{r=1}^{s-1} \frac{g_{i_s,r}\mathbb{I}\left[A_r = i_s\right]}{T_{i_s}(s-1)}}_{\in \{-1,1\}}+ \sigma\sqrt{\underbrace{\frac{1+T_{i_s}(s-1)}{T_{i_s}^2(s-1)}}_{\le 2}\underbrace{\left(1+ 2\log\left(\frac{K(1+T_{i_s}(s-1))^{1/2}}{\delta'}\right)\right)}_{\le 1+2\log(K(1+n)/\delta')}} ,
\end{align*}
and therefore
\begin{align*}
    \lvert \mu_{i_s} - \muit{i_s}{s-1}\rvert \le 2 + \sigma\sqrt{2\left(1+2\log\left(\frac{K(1+n)}{\delta'}\right) \right)} =: \mathcal{P}_n, \qquad \forall s\in\{1,\ldots,n\}.
\end{align*}
      
 Define a martingale $Z_{t-K} : = N_t$ and the martingale difference sequence $Y_s : = Z_s - Z_{s-1}$. Then we have, for any $s \in \{K+1,\ldots,t\}$,
 \begin{align*}
      \lv Y_{s-K} \rv_{op}  = \lv Y_{s-K}\rv_2 \le \left\lv \alpha_{i_s,s}\left( \mu_{i_s} - \muit{i_s}{s-1}\right) \right\rv_{2} 
     & \le \lv\alpha_{i_s,s}\rv_2 \lvert\mu_{i_s} -\muit{i_s}{s-1} \rvert  \le  \lvert\mu_{i_s} -\muit{i_s}{s-1} \rvert \le \mathcal{P}_n.
 \end{align*}
 We also have
\begin{align*} 
\left\lv \mathbb{E}_{s-1}\left[\alpha_{i_s,s}\alpha_{i_s,s}^{\top}\left( \mu_{i_s}-\muit{i_s}{s-1}\right)^2\right]\right\rv_{op} \le \mathcal{P}_n^2 \lv \Sigma_c\rv_{op} \le \mathcal{P}_n^2,
\end{align*}
and
\begin{align*}
\left\lv \mathbb{E}_{s-1}\left[\alpha_{i_s,s}^{\top}\alpha_{i_s,s}\left( \mu_{i_s}-\muit{i_s}{s-1}\right)^2\right]\right\rv_{op} \le \mathcal{P}_n^2 \lv \alpha_{i_s,s}\rv^2_{2} \le \mathcal{P}_n^2.
\end{align*}
Invoking Theorem \ref{thm:matrixfreedman} with $R=\mathcal{P}_n$ and $\omega^2 = \mathcal{P}_n^2 (t-K)$, we get
\begin{align*}
    \mathbb{P}\left\{ \lv N_t \rv_{2} \ge \frac{\mathcal{P}_n}{3}\log\left( \frac{2dn}{\delta'}\right)+ \frac{\mathcal{P}_n}{3}\sqrt{18(t-K)\log\left( \frac{2dn}{\delta'}\right)+\log^2\left( \frac{2dn}{\delta'}\right)}\right\} \le \frac{\delta'}{n}.
\end{align*}
From the definition of $\updela$ in Eq. \eqref{def:up} and applying the union bound over all $t\in \{K+1,\ldots,n\}$, we get
\begin{align*}
    \mathbb{P}\left\{ \exists t \in \{K+1,\ldots,n\}: \lv N_t \rv_{2} \ge \updela\right\} \le \delta'.
\end{align*}
This completes the proof.
\end{Proof}

The following lemma is an analog of Lemma~\ref{lem:csimplebound} under the complex model.

\begin{lemma}
\label{lem:updatedbiasbound}
Under the complex model, with probability at least $1-2\delta'$ we have, for all $i \in [1,\ldots,K]$ and for all $t\in \{K+1,\ldots,n\}$,
\begin{align*}
    \lvert \mu_i - \bar{g}_{i,t} \rvert \le (\sigma + 1) \left[\frac{1+T_i(t)}{T_i^2(t)}\left(1+2\log\left( \frac{K(1+T_i(t))^{\frac{1}{2}}}{\delta'}\right)\right)\right]^{1/2} .
\end{align*}
\end{lemma}
\begin{Proof}
Under the complex model, we know that
\begin{align*}
    \lvert  \mu_i - \bar{g}_{i,t} \rvert & = \left\lvert \frac{\sum_{s=1}^t g_{i,s}\mathbb{I}[A_s=i]}{T_i(t)} -\mu_i\right\rvert\\
    & = \left \lvert  \frac{\sum_{s=1}^t (\mu_i +\eta_{i,s} + \langle \alpha_{i,s},\theta^* \rangle )\mathbb{I}[A_s=i]}{T_i(t)} -\mu_i\right\rvert\\
    & \le \left \lvert  \frac{\sum_{s=1}^t (\mu_i +\eta_{i,s}  )\mathbb{I}[A_s=i]}{T_i(t)} -\mu_i\right\rvert + \left\lvert \frac{\sum_{s=1}^t \langle \alpha_{i,s},\theta^* \rangle \mathbb{I}[A_s=i]}{T_i(t)}\right\rvert . \numberthis \label{e:fixthelemma1forcomplexmodelmastereq}
\end{align*}
By invoking Lemma~6 in \citep{abbasi2011improved} with probability $1-\delta'$ for all $i\in [K]$ and for all $t \ge K+1$ the first term is bounded above by
\begin{align*}
    \left \lvert  \frac{\sum_{s=1}^t (\mu_i +\eta_{i,s} )\mathbb{I}[A_s=i]}{T_i(t)} -\mu_i\right\rvert \le \sigma \left[\frac{1+T_i(t)}{T_i^2(t)}\left(1+2\log\left( \frac{K(1+T_i(t))^{\frac{1}{2}}}{\delta'}\right)\right)\right]^{1/2}.
\end{align*}
Now let us analyze the second term in Eq.~\eqref{e:fixthelemma1forcomplexmodelmastereq},
\begin{align*}
    \left\lvert \frac{\sum_{s=1}^t \langle \alpha_{i,s},\theta^* \rangle \mathbb{I}[A_s=i]}{T_i(t)}\right\rvert & = \frac{ \left\lvert \sum_{s=1}^t \langle \alpha_{i,s},\theta^* \rangle \mathbb{I}[A_s=i]\right\rvert}{T_i(t)}.
\end{align*}
By applying Theorem~\ref{thm:selfnormalized} with $Y_s = \mathbb{I}[A_s=i] $, $\xi_s =\langle \alpha_{i,s},\theta^* \rangle$ and $V =1$ we get that, with probability $1-\delta'/K$, for all $t \ge K+1$,
\begin{align*}
    \frac{\left\lvert \sum_{s=1}^t \langle \alpha_{i,s},\theta^* \rangle  \mathbb{I}[A_s = i]\right\rvert}{T_i(t)}  \le \sqrt{\frac{1+T_i(t)}{T_i(t)^2}\left(1+2\log\left(\frac{K (1+T_i(t))^{1/2}}{\delta'}\right)\right)},
\end{align*}
since $\langle \alpha_{i,s},\theta^*\rangle$ is conditionally $1$-sub-Gaussian. 
A union bound over all arms combined with Eq.~\eqref{e:fixthelemma1forcomplexmodelmastereq} above completes the proof.
\end{Proof}

The next lemma provides a high probability upper bound on the terms $\sum_{s=K+1}^t \langle \alpha_{j_s,s},\tilde{\theta}_s-\theta^*\rangle$ and $\sum_{s=K+1}^t \langle \alpha_{\kappa_s,s},\theta^*-\tilde{\theta}_s\rangle$. Note that in each term the context vectors $\alpha_{j_s,s}$ and $\alpha{\kappa_s,s}$ are not independent of $\tilde{\theta}_s-\theta^*$ which is why we require a careful martingale analysis in the proof.

\begin{lemma}

For all $K+1 \le t \le n$ we have
\begin{align*}
    \sum_{s= K+1}^{t} \langle \alpha_{j_s,s} ,\tilde{\theta}_s - \theta^* \rangle &\le \mathcal{Q}_{\delta'}(t,n), \text{ and}\\
    \sum_{s= K+1}^{t} \langle \alpha_{\kappa_s,s} ,\theta^*-\tilde{\theta}_s \rangle & \le \mathcal{Q}_{\delta'}(t,n)
\end{align*}
with probability at least $1-14\delta' n$.
\label{l:concentrationofalphatimestheta}
\end{lemma}
\begin{Proof} 
We define the filtration $\mathcal{G}_{s-1}$ that is the sigma-algebra generated by the random variables $ \{A_{s'},\alpha_{A_{s'},s'}\}_{s'=1}^{s-1}$.
Note that this filtration is the $\sigma$-algebra born out of the algorithmic actions and contexts seen up until the end of round $s-1$.

First we will establish an upper bound on 
\begin{align*}
    W_t : = \sum_{s=K+1}^{t} \langle \alpha_{j_s,s},\tilde{\theta}_s -\theta^*\rangle.
\end{align*}
Define $Z_s : = \max_{\ell \in [K]} \langle \alpha_{\ell,s} ,\tilde{\theta}_s - \theta^* \rangle -  \mathbb{E}\left[ \max_{\ell \in [K]} \langle \alpha_{\ell,s},\tilde{\theta}_s - \theta^* \rangle \Big| \mathcal{G}_{s-1}\right]$. Finally, let,
\begin{align*}
    Y_t : = \sum_{s=K+1}^{t} Z_s = \sum_{s= K+1}^t \left(\max_{\ell \in [K]} \langle \alpha_{\ell,s},\tilde{\theta}_s - \theta^* \rangle - \mathbb{E}\left[ \max_{\ell \in [K]} \langle \alpha_{\ell,s},\tilde{\theta}_s - \theta^* \rangle \Big| \mathcal{G}_{s-1}\right]\right).
\end{align*}

Clearly we have that for all $t$,
\begin{align}
   \nonumber  W_t &\le Y_t+  \sum_{s=K+1}^t  \mathbb{E}\left[ \max_{\ell \in [K]} \langle \alpha_{\ell,s},\tilde{\theta}_s - \theta^* \rangle \Big| \mathcal{G}_{s-1}\right] \\&= \sum_{s= K+1}^t \max_{\ell \in [K]} \langle \alpha_{\ell,s},\tilde{\theta}_s - \theta^* \rangle . \label{eq:upperboundmartingale}
\end{align}
We will now proceed to prove a high probability upper bound on $$\sum_{s=K+1}^{t}  \max_{\ell \in [K]} \langle \alpha_{\ell,s},\tilde{\theta}_s - \theta^* \rangle,$$ which also provides a bound on $W_t$. 

Let the sum of the conditional variances be
\begin{align}
    \mathcal{V}_t : = \sum_{s=K+1}^t \mathbb{E}\left[Z_s^2\Big| \mathcal{G}_{s-1}\right]. \label{def:variance}
\end{align}
Each term $\lvert Z_s \rvert \le 4$, since $\lv \alpha_{\ell,s} \rv_2 \le 1$ and $\lv \tilde{\theta}_s - \theta^* \rv_2 \le \sqrt{2}$ (since both $\lv\tilde{\theta}_s\rv_2, \lv \theta^* \rv_2 \le 1$). Therefore, by Freedman's inequality \citep[see, e.g.,][Lemma~1]{peel2013empirical} applied to the martingale $Y_t$,
\begin{align*}
 \mathbb{P}\left\{ \left( Y_t \ge  u \right) \cap \left(\mathcal{V}_t \le \omega^2\right) \right\}
    \le \exp\left(-\frac{u^2}{2\omega^2+8u/3}\right). 
\end{align*}

Note that
\begin{align*}
 \mathbb{P}\left[ Y_t \ge  u \right]   & = \mathbb{P}\left[  Y_t \ge  u | \mathcal{V}_t \le \omega^2\right] \mathbb{P}\left[\mathcal{V}_t \le \omega^2\right] + \mathbb{P}\left[Y_t \ge  u | \mathcal{V}_t \ge \omega^2\right] \mathbb{P}\left[\mathcal{V}_t \ge \omega^2\right] \\
  & \le \mathbb{P}\left\{\left( Y_t \ge  u \right) \cap \left(\mathcal{V}_t \le \omega^2\right) \right\} + \mathbb{P}\left[\mathcal{V}_t \ge \omega^2\right]\\
  & \le \exp\left(-\frac{u^2}{2\omega^2+8u/3}\right) + \mathbb{P}\left[\mathcal{V}_t \ge \omega^2\right]. \label{eq:ytupperbound} \numberthis
\end{align*}

Combining this upper bound on the sum of the conditional variance in Lemma~\ref{l:variancebound} along with the setting of $\omega^2 = 64 \log(K) \rho_{\max} \sum_{s=K+1}^{t} \kenvs^2$ we obtain
\begin{align*}
    \mathbb{P}\left[Y_t \ge u \right] \le \exp\left(-\frac{u^2}{128 \log(K) \rho_{\max} \sum_{s=K+1}^{t} \kenvs^2 + 8u/3}\right) + 3\delta'.
\end{align*}
Choosing $$u_t^* = 16 \sqrt{\log(K) \rho_{\max} \left(\sum_{s=K+1}^t \kenvs^2\right) \log(1/\delta')} + \frac{8}{3}\log(1/\delta'),$$
ensures that
\begin{align*}
    \mathbb{P}\left[Y_t \ge u_t^* \right] \le 4\delta'.
\end{align*}
By the definition of $Y_t$ we know that
\begin{align*}
    &\mathbb{P}\left[\sum_{s=K+1}^t \max_{\ell,s}\langle \alpha_{\ell,s},\tilde{\theta}_s - \theta^*\rangle  \le 4\sqrt{2} \sqrt{\log(K) \rho_{\max}} \sum_{s=K+1}^{t} \kenvs + u_t^* \right] \\& \ge \mathbb{P}\left[\left(\sum_{s=K+1}^t \mathbb{E}\left[\max_{\ell \in [K]} \langle \alpha_{\ell,s},\tilde{\theta}_s - \theta^* \rangle \Big| \mathcal{G}_{s-1}\right] \le 4\sqrt{2} \sqrt{\log(K) \rho_{\max}} \sum_{s=K+1}^{t} \kenvs \right) {\mathlarger{\mathlarger\cap}} \left( Y_t \le u_t^*\right)\right].
\end{align*}
Therefore by again invoking Lemma~\ref{l:variancebound},
\begin{align*}
    &\mathbb{P}\left[\sum_{s=K+1}^t \max_{\ell,s}\langle \alpha_{\ell,s},\tilde{\theta}_s - \theta^*\rangle  \ge 4\sqrt{2} \sqrt{\log(K) \rho_{\max}} \sum_{s=K+1}^{t} \kenvs + u_t^* \right] \\& \le \mathbb{P}\left[\left(\sum_{s=K+1}^t \mathbb{E}\left[\max_{\ell \in [K]} \langle \alpha_{\ell,s},\tilde{\theta}_s - \theta^* \rangle \Big| \mathcal{G}_{s-1}\right] \ge 4\sqrt{2} \sqrt{\log(K) \rho_{\max}} \sum_{s=K+1}^{t} \kenvs \right) {\mathlarger{\mathlarger\cup}} \left( Y_t \ge u_t^*\right)\right] \\
    &\le \mathbb{P}\left[\sum_{s=K+1}^t \mathbb{E}\left[\max_{\ell \in [K]} \langle \alpha_{\ell,s},\tilde{\theta}_s - \theta^* \rangle \Big| \mathcal{G}_{s-1} \right] \ge 4\sqrt{2} \sqrt{\log(K) \rho_{\max}} \sum_{s=K+1}^{t} \kenvs\right] +  \mathbb{P}\left[Y_t \ge u_t^* \right] \\
    & \le 7\delta'.
\end{align*}
We take a union bound over the rounds $t \in \{K+1,\ldots,n\}$ to infer that
\begin{align*}
    &\mathbb{P}\left[\exists t: \sum_{s=K+1}^t \max_{\ell,s}\langle \alpha_{\ell,s},\tilde{\theta}_s - \theta^*\rangle  \ge 4\sqrt{2} \sqrt{\log(K) \rho_{\max}} \sum_{s=K+1}^{t} \kenvs + u_t^* \right] \le 7\delta' n.
\end{align*}
The definition of $\mathcal{Q}_{\delta'}(t,n)$ along with inequality~\eqref{eq:upperboundmartingale} completes the proof of the first part. 
By applying an identical argument we can also establish the second part of the lemma with probability at least $1-7\delta'n$. We finish the proof by taking a union bound such that both parts of the lemma simultaneously hold. 
\end{Proof}

\begin{lemma}\label{l:variancebound} We borrow all notation from the previous lemma. For any $t \in \{K+1,\ldots,t\}$,
\begin{align*}
    \mathbb{P}\left[\mathcal{V}_t \ge 64 \log(K) \rho_{\max} \sum_{s=K+1}^{t} \kenvs^2\right] \le 3\delta',
\end{align*}
and,
\begin{align*}
    \mathbb{P}\left[\sum_{s=K+1}^t \mathbb{E}\left[\max_{\ell \in [K]} \langle \alpha_{\ell,s},\tilde{\theta}_s - \theta^* \rangle \Big| \mathcal{G}_{s-1}\right] \ge 4\sqrt{2} \sqrt{\log(K) \rho_{\max}} \sum_{s=K+1}^{t} \kenvs\right] \le 3\delta'.
\end{align*}
\end{lemma}
\begin{Proof}
For any $\lambda \in \mathbb{R}$, we obtain the following chain of inequalities:
\begin{align*}
    &\mathbb{E}\left[Z_s^2 |\mathcal{G}_{s-1}\right] \\ &= \frac{2}{\lambda}\log\left(\exp\left(\frac{\lambda}{2} \mathbb{E}\left[Z_s^2 |\mathcal{G}_{s-1}\right]\right)\right)\\ 
    & = \frac{2}{\lambda}\log\left(\exp\left(\frac{\lambda}{2} \mathbb{E}\left[\left(\max_{\ell \in [K]} \langle \alpha_{\ell,s} ,\tilde{\theta}_s - \theta^* \rangle -  \mathbb{E}\left[ \max_{\ell \in [K]} \langle \alpha_{\ell,s},\tilde{\theta}_s - \theta^* \rangle \Big| \mathcal{G}_{s-1}\right]\right)^2 \Big|\mathcal{G}_{s-1}\right]\right)\right) \\
    & \overset{(i)}{\le} \frac{2}{\lambda}\log\left(\mathbb{E}\left[\exp\left(\frac{\lambda}{2} \left(\max_{\ell \in [K]} \langle \alpha_{\ell,s} ,\tilde{\theta}_s - \theta^* \rangle -  \mathbb{E}\left[ \max_{\ell \in [K]} \langle \alpha_{\ell,s},\tilde{\theta}_s - \theta^* \rangle \Big| \mathcal{G}_{s-1}\right]\right)^2 \right)\Big|\mathcal{G}_{s-1}\right]\right) \\
      & = \frac{2}{\lambda}\log\left(\mathbb{E}\left[\max_{\ell \in [K]}\exp\left(\frac{\lambda}{2} \left(\langle \alpha_{\ell,s} ,\tilde{\theta}_s - \theta^* \rangle -  \mathbb{E}\left[ \max_{\ell \in [K]} \langle \alpha_{\ell,s},\tilde{\theta}_s - \theta^* \rangle \Big| \mathcal{G}_{s-1}\right]\right)^2 \right)\Big|\mathcal{G}_{s-1}\right]\right) \\
    & \le \frac{2}{\lambda}\log\left(\mathbb{E}\left[\sum_{\ell \in [K]}\exp\left(\frac{\lambda}{2} \left( \langle \alpha_{\ell,s} ,\tilde{\theta}_s - \theta^* \rangle -  \mathbb{E}\left[ \max_{\ell \in [K]} \langle \alpha_{\ell,s},\tilde{\theta}_s - \theta^* \rangle \Big| \mathcal{G}_{s-1}\right]\right)^2 \right)\Big|\mathcal{G}_{s-1}\right]\right) \\
    & \le \frac{2}{\lambda}\log\left(\mathbb{E}\left[\sum_{\ell \in [K]}\exp\left(\lambda \left( \langle \alpha_{\ell,s} ,\tilde{\theta}_s - \theta^* \rangle\right)^2 + \lambda \left(\mathbb{E}\left[ \max_{\ell \in [K]} \langle \alpha_{\ell,s},\tilde{\theta}_s - \theta^* \rangle \Big| \mathcal{G}_{s-1}\right]\right)^2 \right)\Big|\mathcal{G}_{s-1}\right]\right) \\
    & = \frac{2}{\lambda}\log\left(e^{\lambda\left(\mathbb{E}\left[ \max_{\ell \in [K]} \langle \alpha_{\ell,s},\tilde{\theta}_s - \theta^* \rangle \Big| \mathcal{G}_{s-1}\right]\right)^2}\mathbb{E}\left[\sum_{\ell \in [K]}\exp\left(\lambda \left( \langle \alpha_{\ell,s} ,\tilde{\theta}_s - \theta^* \rangle\right)^2  \Big|\mathcal{G}_{s-1}\right)\right]\right) \\
    & \le \frac{2}{\lambda}\log\left(Ke^{\lambda\left(\mathbb{E}\left[ \max_{\ell \in [K]} \langle \alpha_{\ell,s},\tilde{\theta}_s - \theta^* \rangle \Big| \mathcal{G}_{s-1}\right]\right)^2}\mathbb{E}\left[\exp\left(\lambda \left( \langle \alpha_{\ell,s} ,\tilde{\theta}_s - \theta^* \rangle\right)^2  \Big|\mathcal{G}_{s-1}\right)\right]\right),
\end{align*}
where $(i)$ follows by Jensen's inequality. Now since conditioned on $\mathcal{G}_{s-1}$, the random variable $\langle \alpha_{\ell,s},\tilde{\theta}_s - \theta^* \rangle$ is $\rho_{\max} \lv \tilde{\theta}_s - \theta^* \rv^2$-sub-Gaussian, the random variable, $\langle \alpha_{\ell,s},\tilde{\theta}_s - \theta^* \rangle^2$ is sub-exponential \citep[see, e.g.,][Lemma~2.7.6]{vershynin2018high}, and therefore, we get:
\begin{align*}
    \mathbb{E}\left[\exp\left(\lambda \left( \langle \alpha_{\ell,s} ,\tilde{\theta}_s - \theta^* \rangle\right)^2  \Big|\mathcal{G}_{s-1}\right)\right] \le \exp(\lambda^2 \rho_{\max}^2\lv \tilde{\theta}_s - \theta^* \rv^4 ),
\end{align*}
for all $\lvert\lambda\rvert \le \frac{1}{\sqrt{2}\rho_{\max} \lv \tilde{\theta}_s - \theta^* \rv^2}$. Substituting this above, with the choice of $\lambda = \frac{1}{\sqrt{2}\rho_{\max} \lv \tilde{\theta}_s - \theta^* \rv^2}$ leads to the upper bound
\begin{align}
     \mathbb{E}\left[Z_s^2 |\mathcal{G}_{s-1}\right]  \le 4\sqrt{2} \log(K) \rho_{\max} \lv \tilde{\theta}_s - \theta^* \rv^2 + 2\left(\mathbb{E}\left[ \max_{\ell \in [K]} \langle \alpha_{\ell,s},\tilde{\theta}_s - \theta^* \rangle \Big| \mathcal{G}_{s-1}\right]\right)^2. \label{eq:variance_upper_bound}
\end{align}

Next, we bound the expected value as follows,
\begin{align}
    \mathbb{E}\left[\max_{\ell \in [K]} \langle\alpha_{\ell,s},\tilde{\theta}_s -\theta^* \rangle \Big| \mathcal{G}_{s-1}\right] & \le 2\sqrt{2 \rho_{\max}\log(K)} \lv \tilde{\theta}_s - \theta^* \rv_2,  \label{eq:maximumexpectedvaluebound}
\end{align}
where the upper bound follows since $\alpha_{k,s}$ is a $\rho_{\max}$-sub-Gaussian random vector for every $k \in [K]$ and $\tilde{\theta}_s-\theta^*$ is $\mathcal{G}_{s-1}$-measurable \citep[see, e.g.,][Exercise~2.12]{wainwright2019high}. Combining this with inequality~\eqref{eq:variance_upper_bound},
\begin{align*}
    \mathbb{E}\left[Z_s^2 |\mathcal{G}_{s-1}\right]  & \le 16 \log(K) \rho_{\max} \lv \tilde{\theta}_s - \theta^* \rv^2.
\end{align*}
Further, by Lemma~\ref{prop:thetaconfidence} we know that
\begin{align}
    \mathbb{P}\left[\exists s:  \lv \tilde{\theta}_s - \theta^* \rv \ge 2\mathcal{K}_{\delta'}(s-1,n)\right] \le 3\delta'. \label{eq:thetaupperboundhighprobability}
\end{align}
Therefore,
\begin{align*}
    \mathbb{P}\left[\mathcal{V}_t \ge 64 \log(K) \rho_{\max} \sum_{s=K+1}^{t} \kenvs^2\right] \le 3\delta'.
\end{align*}
This proves the first part of the lemma. For the second part, notice that combining the inequalities~\eqref{eq:maximumexpectedvaluebound} and \eqref{eq:thetaupperboundhighprobability} gives us the desired bound,
\begin{align*}
    \mathbb{P}\left[\sum_{s=K+1}^t \mathbb{E}\left[\max_{\ell \in [K]} \langle \alpha_{\ell,s},\tilde{\theta}_s - \theta^* \rangle \right] \ge 4\sqrt{2} \sqrt{\log(K) \rho_{\max}} \sum_{s=K+1}^{t} \kenvs\right] \le 3\delta'.
\end{align*}
\end{Proof}
\section{Concentration Inequalities and Technical Results}  \label{sec:tech}

In this section we state technical concentration inequalities that are useful in our proofs.
We start by defining notation specific to this section.

Let $\{\mathcal{H}_{t}\}_{t=0}^{\infty}$ be a filtration. Let $\{\xi_t\}_{t=1}^{\infty}$ be a real-valued stochastic process such that $\xi_t$ is $\mathcal{H}_{t}$-measurable and $\xi_t$ is conditionally $\sigma$-sub-Gaussian. Let $\{Y_t\}_{t=1}^\infty$ be an $\mathbb{R}^d$-valued stochastic process such that $Y_t$ is $\mathcal{H}_{t-1}$-measurable. Assume that $V$ is a $d\times d$ positive definite matrix. For any $t>0$ define
\begin{align*}
    V_t := V + \sum_{s=1}^t Y_s Y_s^{\top}, \qquad S_t := \sum_{s=1}^{t} \xi_s Y_s.
\end{align*}
With this setup in place, the following is a re-statement of Theorem 1 of \citet{abbasi2011improved}, which is essentially a self-normalized concentration inequality.

\begin{theorem}\label{thm:selfnormalized} For any $\delta'>0$, we have
\begin{align*}
    S_t^{\top}V_t^{-1}S_t = \lv S_t \rv_{V_t^{-1}}^2 \le 2\sigma^2 \log\left( \frac{\det(V_t)^{1/2}\det(V)^{-1/2}}{\delta'}\right)
\end{align*}
with probability at least $1 - \delta'$ for all $t \ge 0$.
\end{theorem}

Next we state a version of the Matrix Freedman Inequality due to \citet[Corollary 1.3]{tropp2011freedman} that we use multiple times in our arguments. 
For a filtration $\{\mathcal{H}_s\}_{s \geq 1}$, a matrix martingale is defined as a sequence $\{Z_s: s=0,1,\ldots\}$ such that $Z_0 = 0$ and 
\begin{align*}
    \mathbb{E}\left[Z_s \lvert \mathcal{H}_{s-1}\right] = Z_{s-1} \qquad \text{and } \qquad \mathbb{E}\left[\lv Z_s\rv_{op}\right] \le \infty, \qquad \text{for } s=1,\ldots.
\end{align*}
Also define the martingale difference sequence $X_s := Z_s - Z_{s-1}$. 

\begin{theorem}\label{thm:matrixfreedman}Consider a matrix martingale $\{Z_s:s = 0,1,\ldots\}$ whose values are matrices with dimension $d_1 \times d_2$, and let $\{X_s:s=0,1,\ldots\}$ be the martingale difference sequence. Assume that the difference sequence is almost surely uniformly bounded, that is,
\begin{align*}
    \lv X_s\rv_{op} \le R \qquad \text{a.s. } \qquad \text{for }s = 1,2\ldots
\end{align*}
Define two predictable quadratic variation processes of the martingale:
\begin{align*}
    W_{col,t} &: = \sum_{s=1}^t \mathbb{E}\left[X_sX_s^{\top} \lvert \mathcal{H}_{s-1}\right] \qquad \text{and}\\
    W_{row,t} &: = \sum_{s=1}^t \mathbb{E}\left[X_s^{\top}X_s \lvert \mathcal{H}_{s-1}\right] \qquad \text{for }t=1,2,\ldots
\end{align*}
Then for all $u\ge 0$ and $\omega^2 >0$, we have
\begin{align*}
    &\mathbb{P}\left\{ \exists t \ge 0: \lv Z_t\rv_{op}  \ge u  \text{ and } \max\left\{\lv W_{col,t}\rv_{op},\lv W_{row,t}\rv_{op} \right\}\le \omega^2 \right\} \\ &\qquad \qquad \qquad \qquad \qquad \qquad \qquad \qquad \qquad \qquad \le (d_1+d_2)\exp\left(-\frac{u^2/2}{\omega^2+Ru/3}\right).
\end{align*}

\end{theorem}

The final technical result we recap characterizes the growth of the determinant of the matrix $V_n$, and is useful in constructing our confidence sets for the estimate of $\thetastar$.
This result is a restatement of Lemma 19.1 in the pre-print \citep{lattimore2018bandit}.

\begin{lemma}\label{lem:detcontrol} Let $V_0\in \mathbb{R}^{d\times d}$ be a positive definite matrix and $z_1,\ldots,z_n \in \mathbb{R}^d$ be a sequence of vectors with $\lv z_t \rv_2 \le L < \infty$ for all $t\in [n]$. 
Further, let $v_0 := \tr(V_0)$ and $V_n := V_0 +\sum_{s=1}^n z_sz_s^{\top}$.
Then, we have
\begin{align*}
    \log\left(\frac{\det(V_n)}{\det(V_0)}\right) \le d\log\left( \frac{v_0 +nL^2}{d\det^{1/d}(V_0)}\right).
\end{align*}
\end{lemma}
\printbibliography
\end{document}